\documentclass{article}

\usepackage[nonatbib,final]{neurips_2021}

\usepackage[utf8]{inputenc} 
\usepackage[T1]{fontenc}    
\usepackage{hyperref}       
\usepackage{url}            
\usepackage{booktabs}       
\usepackage{amsfonts}       
\usepackage{nicefrac}       
\usepackage{microtype}      
\usepackage{xcolor}         

\usepackage[pdftex]{graphicx}
\usepackage{enumitem}

\usepackage{algorithm}
\usepackage{algorithmic}
\usepackage{subcaption}
\usepackage{wrapfig}
\usepackage{multirow}
\usepackage{sidecap}

\newcommand{\dataonly}{\textsc{TaskOnly}}
\newcommand{\taskonly}{\textsc{TaskOnly}}
\newcommand{\ruleonly}{\textsc{RuleOnly}}
\newcommand{\taskwithrule}{\textsc{Task\&Rule}}
\newcommand{\ours}{\textsc{DeepCTRL}}

\newcommand{\betadist}{\text{Beta}}

\newcommand{\negone}{\textsc{Neg0.1}}
\newcommand{\negtwo}{\textsc{Neg0.2}}
\newcommand{\negthree}{\textsc{Neg0.3}}
\newcommand{\postwo}{\textsc{Pos0.2}}
\newcommand{\posthree}{\textsc{Pos0.3}}

\usepackage{amsmath,amsfonts,bm}

\def\eqref#1{equation~\ref{#1}}

\def\1{\bm{1}}

\def\vx{{\bm{x}}}
\def\vy{{\bm{y}}}
\def\vz{{\bm{z}}}

\DeclareMathAlphabet{\mathsfit}{\encodingdefault}{\sfdefault}{m}{sl}
\SetMathAlphabet{\mathsfit}{bold}{\encodingdefault}{\sfdefault}{bx}{n}

\def\gD{{\mathcal{D}}}

\def\gL{{\mathcal{L}}}

\def\gN{{\mathcal{N}}}

\def\sR{{\mathbb{R}}}

\title{Controlling Neural Networks with Rule Representations}

\author{%
  Sungyong Seo,\ \ Sercan \"{O}. Ar{\i}k,\ \ Jinsung Yoon,\ \ Xiang Zhang,\ \ Kihyuk Sohn,\ \ Tomas Pfister \\
  Google Cloud AI \\
  Sunnyvale, CA, USA \\
  \texttt{\{sungyongs,soarik,jinsungyoon,fancyzhx,kihyuks,tpfister\}@google.com}
}

\begin{document}

\maketitle

\begin{abstract}

We propose a novel training method that integrates rules into deep learning, in a way the strengths of the rules are controllable at inference.
Deep Neural Networks with Controllable Rule Representations ({\ours}) incorporates a rule encoder into the model coupled with a rule-based objective, enabling a shared representation for decision making.
{\ours} is agnostic to data type and model architecture. 
It can be applied to any kind of rule defined for inputs and outputs.
The key aspect of {\ours} is that it does not require retraining to adapt the rule strength -- at inference, the user can adjust it based on the desired operation point on accuracy vs. rule verification ratio.
In real-world domains where incorporating rules is critical -- such as Physics, Retail and Healthcare -- we show the effectiveness of {\ours} in teaching rules for deep learning. 
{\ours} improves the trust and reliability of the trained models by significantly increasing their rule verification ratio, while also providing accuracy gains at downstream tasks. 
Additionally, {\ours} enables novel use cases such as hypothesis testing of the rules on data samples, and unsupervised adaptation based on shared rules between datasets.

\end{abstract}

\section{Introduction}\label{sec:intro}

\begin{figure}[!htbp]
    \centering
    \includegraphics[width=0.7\textwidth]{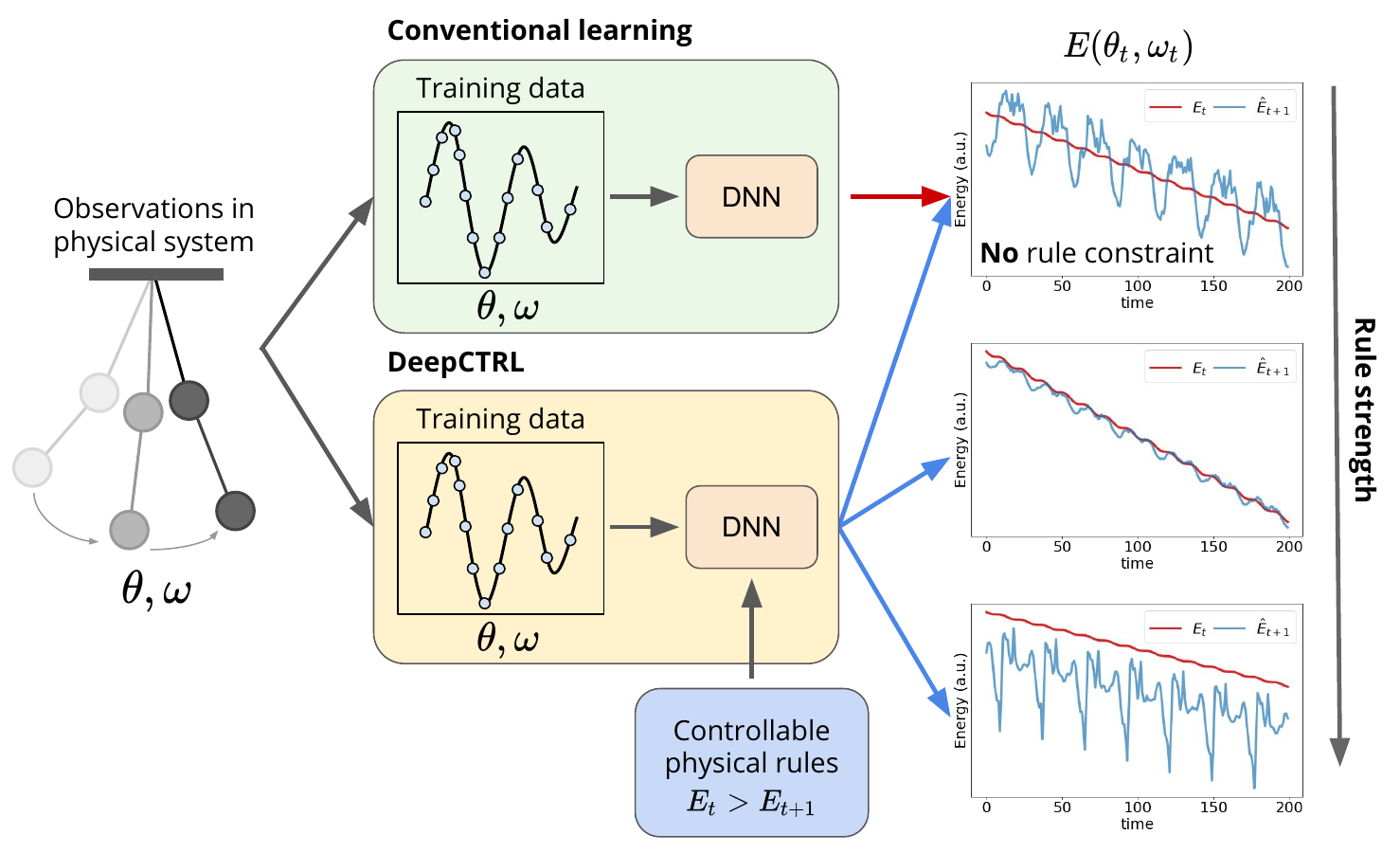}
    \caption{Overview of {\ours}. 
    When a DNN is trained only with the task-specific objective (in this example, predicting next positions/velocities of two objects connected in a pendulum), it may easily violate the rule ($E_t>E_{t+1}$) that it must have followed according to the energy damping rule from physics (top graph).
    {\ours} (with outputs shown via blue arrows) provides a controllable mechanism that enables the rule dependency to be adjusted at inference time in order to achieve an optimal behavior (middle graph) in regards to accuracy and rule verification ratio. With increased rule strength, {\ours} yields an operation point (bottom graph) where 
 is satisfied for all time steps.
    } \label{fig:dp-motivation}
\end{figure}

Deep neural networks (DNNs) excel at numerous tasks such as image classification~\cite{szegedy2016rethinking,touvron2019fixing}, machine translation~\cite{liu2020very,vaswani2017attention}, time series forecasting~\cite{eisenach2020mqtransformer,tft}, and tabular learning~\cite{tabnet, node}.
DNNs get more accurate as the size and coverage of training data increase \cite{hestness2017deep}. 
While investing in high-quality and large-scale labeled data is one path, another is utilizing prior knowledge -- concisely referred to as `rules': reasoning heuristics, equations, associative logic, constraints or blacklists.
In most scenarios, labeled datasets are not sufficient to teach all rules present about a task \cite{asseman2018learning, fioretto2020lagrangian, lutter2019deep,pmlr-v84-narasimhan18a}.
Let us consider an example from Physics: the task of predicting the next state in a double pendulum system, visualized in Fig.~\ref{fig:dp-motivation}. 
Although a `data-driven' black-box model, fitted with conventional supervised learning, can fit a relatively accurate mapping from the current state to next, it can easily fail to capture the canonical rule of `energy conservation'. 
In this work, we focus on how to teach `rules' in effective ways so that DNNs absorb the knowledge from them in addition to learning from the data for the downstream task.

The benefits of learning from rules are multifaceted. 
First, rules can provide extra information for cases with minimal data supervision, improving the test accuracy. 
Second, the rules can improve trust and reliability of DNNs.
One major bottleneck for widespread use of DNNs is them being `black-box'.  
The lack of understanding of the rationale behind their reasoning and inconsistencies of their outputs with human judgement often reduce the trust of the users ~\cite{arya2019one,ribeiro2016should}.
By incorporating rules, such inconsistencies can be minimized and the users' trust can be improved. 
For example, if a DNN for loan delinquency prediction can absorb all the decision heuristics used at a bank, the loan officers of the bank can rely on the predictions more comfortably.
Third, DNNs are sensitive to slight changes to the inputs that are human-imperceptible~\cite{goodfellow2014explaining,kurakin2016adversarial,yuan2019adversarial}.
With rules, the impact of these changes can be minimized as the model search space is further constrained to reduce underspecification ~\cite{damour2020underspecification,doshi2017towards}.

When `data-driven' and `rule-driven' learning are considered jointly, a fundamental question is how to balance the contribution from each. 
Even when a rule is known to hold 100\% of the time (such as the principles in natural sciences), the contribution of rule-driven learning should not be increased arbitrarily. 
There is an optimal trade-off that depends not only on the dataset, but also on each sample. 
If there are training samples that are very similar to a particular test sample, a weaker rule-driven contribution would be desirable at inference. 
On the other hand, if the rule is known to hold for only a subset of samples (e.g. in Retail, the varying impact of a price change on different products \cite{Browning1986MicroeconomicTA}), the strength of the rule-driven learning contribution should reflect that. 
Thus, a framework where the contributions of data- and rule-driven learning can be controlled would be valuable. 
Ideally, such control should be enabled at inference without the need for retraining in order to minimize the computational cost, shorten the deployment time, and to adjust to different samples or changing distributions flexibly. 


In this paper, we propose {\ours} that enables joint learning from labeled data and rules. 
{\ours} employs separate encoders for data and rules with the outputs combined stochastically to cover intermediate representations coupling with corresponding objectives. 
This representation learning is the key to controllability, as it allows increasing/decreasing the rule strength gradually at inference without retraining.
To convert any non-differentiable rules into differentiable objectives, we propose a novel perturbation-based method.
{\ours} is agnostic to the data type or the model architecture, and {\ours} can be flexibly used in different tasks and with different rules.
We demonstrate {\ours} on important use cases from Physics, Retail, and Healthcare, and show that it: (i) improves the rule verification ratio significantly while yielding better accuracy by merely changing the rule strength at inference; (ii) enables hypotheses to be examined for each sample based on the optimal ratio of rule strength without retraining (for the first time in literature, to the best of our knowledge); and (iii) improves target task performance by changing the rule strength, a desired capability when different subsets of the data are known to satisfy rules with different strengths.


\textbf{Related Work:}
Various methods have been studied to incorporate `rules' into deep learning, considering prior knowledge in wide range of applications.
Posterior regularization \cite{posteriorreg} is one approach to inject rules into predictions.
In \cite{harnessdnnlogicrules}, a teacher-student framework is used, where the teacher network is obtained by projecting the student network to a (logic) rule-regularized subspace and the student network is updated to balance between emulating the teacher’s output and predicting the labels. 
Adversarial learning is utilized in \cite{adversarialdebiasing}, specifically for bias rules, to penalize unwanted biases.
In \cite{fioretto2020lagrangian} a framework is proposed that exploits Lagrangian duality to train with rules. 
In \cite{pmlr-v84-narasimhan18a} learning with constraints is studied, via formulation over the space of confusion matrices and optimization solvers that operate through a sequence of linear minimization steps.
In \cite{shao2020controlvae} constraints are injected via KL divergence for variational autoencoders, for controlling output diversity or disentangled latent factor representations.
{\ours} differentiate from the all aforementioned in how it injects the rules, with the aspect that it allows controllability of the rule strength at the inference without retraining, enabled by accurate learning of the rule representations in the data manifold. This unlocks new capabilities, beyond simply improving rule verification for a target accuracy.

\begin{figure*}[t]
    \centering
\includegraphics[width=0.8\linewidth]{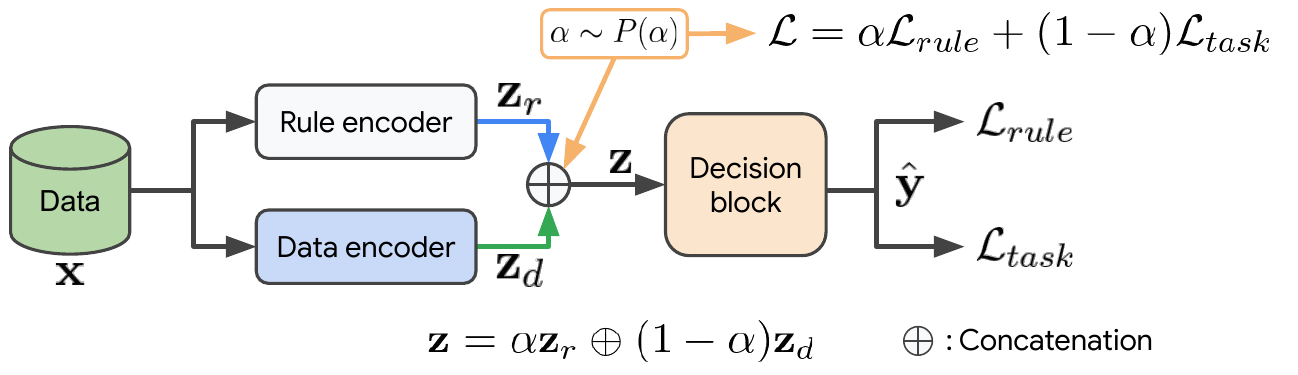}
    \caption{{\ours} for controllable incorporation of a rule within the learning process.
{\ours} introduces two passages for the input-output relationship, a data encoder and rule encoder, that produce two latent representations $\vz_r$ and $\vz_d$. 
These representations are stochastically concatenated with a control parameter $\alpha$ into a single representation $\vz$.
$\vz$ is then fed into a decision block with objectives for each representations, $\mathcal{L}_{rule}$ and $\mathcal{L}_{task}$, again weighed by the control parameter $\alpha$.
$\alpha$ is randomly sampled during training and set by users at inference to adjust the rule strength.}
    \label{fig:method}
\end{figure*}

\section{Learning Jointly from Rules and Task}
\label{sec:model}
\vspace{-0.05in}

Let us consider the conventional approach~\cite{8260755,fioretto2020lagrangian,Goodfellow-et-al-2016} for incorporating rules via combining the training objectives for the supervised task and a term that denotes the violation of rule:
\begin{align}
    \gL=\gL_{task} + \lambda\gL_{rule},\label{eq:convention}
\end{align}
where $\lambda$ is the coefficient for the rule-based objective. There are three limitations of this approach that we aim to address: (i) $\lambda$ needs to be defined before learning (e.g. can be a hyperparameter guided with validation score), (ii) $\lambda$ is not adaptable to target data at inference if there is any mismatch with the training setup, and (iii) $\gL_{rule}$ needs to be differentiable with respect to learnable parameters. 

{\ours} modifies canonical training by creating rule representations, coupled with data representations, which is the key to enable the rule strength to be controlled at inference time.
Fig.~\ref{fig:method} overviews {\ours} and Algorithm~\ref{alg} describes its corresponding training process.
We propose to modify the canonical training approach by introducing \textit{two passages} for the input-output relationship of the task, via \textit{data encoder} $\phi_d$ and \textit{rule encoder} $\phi_r$. 
In this way, our goal is to make each encoder individually learn the latent representations ($\vz_d$ and $\vz_r$), corresponding to extracted information from the labeled data and the rule. 
Then, the two representations are stochastically concatenated (with the operation denoted as $\oplus$) to obtain a single representation $\vz$.
To adjust the relative contributions of data vs. rule encoding, we use a random variable $\alpha\in[0,1]$, which also couples $(\vz_d,\vz_r)$ with the corresponding objectives $(\gL_{task},\gL_{rule})$ (Lines 4 \& 5 in Algorithm~\ref{alg}). 
$\alpha$ is sampled from the distribution $P(\alpha)$. 
The motivation to use a random $\alpha$ is to encourage learning the mapping with a range of values, so that at inference, the model can yield high performance with any particular chosen value. 
The output of the decision block ($\hat{\vy}$) is used in the entire objective.


By modifying the control parameter $\alpha$ at inference, users can control the behavior of the model to adapt it to unseen data. 
The strength of the rule on the output decision can be enhanced by increasing the value of $\alpha$. 
Setting $\alpha=0$ would minimize the contribution of the rule at inference, but as shown in the experiments, the result can still be better than in conventional training since during the training process a wide range of $\alpha$ are considered. 
Typically, an intermediate $\alpha$ value yields the optimal solution given specific performance, transparency and reliability goals.
To ensure that a model shows distinct and robust behavior when $\alpha\rightarrow0$ or $\alpha\rightarrow1$ and thus interpolates accurately later, we propose to sample $\alpha$ more heavily at the two extremes rather than uniformly from $[0,1]$. 
To this end, we choose to sample from a Beta distribution ($\betadist(\beta,\beta)$). 
We observe strong results with $\beta=0.1$ in most cases and in Section 5, we further study the impact of the selection of the prior for $\alpha$.
Since a Beta distribution is highly polarized, each encoder is encouraged to learn distinct representations associated with the corresponding encoder rather than mixed representations.
Similar sampling ideas were also considered in \cite{arik2020protoattend,zhang2018mixup} to effectively sample the mixing weights for representation learning.

\begin{algorithm}[t]
  \caption{Training process for {\ours}.}
   \label{alg}
   \begin{flushleft}
    \textbf{Input}: Training data $\gD=\{(\vx_i,\vy_i):i=1,\cdots,N\}$.\\
    \textbf{Output}: Optimized parameters\\
    \textbf{Require}: Rule encoder $\phi_r$, data encoder $\phi_d$, decision block $\phi$, and distribution $P(\alpha)$.\\
   \end{flushleft}
   \begin{algorithmic}[1]
   \STATE Initialize $\phi_r,\phi_d$, and $\phi$
   \WHILE{not converged}
   \STATE{Get mini-batch $\gD_b$ from $\gD$ and $\alpha_b\in\sR$ from $P(\alpha)$}
   \STATE{Get $\vz=\alpha_b\vz_r\oplus(1-\alpha_b\vz_d)$ where $\vz_r=\phi_r(\gD_b)$ and $\vz_d=\phi_d(\gD_b)$.}
   \STATE{Get $\hat{\vy}=\phi(\vz)$ and compute $\gL=E_{\alpha \sim P(\alpha)} [\alpha\gL_{rule} + \rho(1-\alpha)\gL_{task}]$ where $\rho=\gL_{rule,0}/\gL_{task,0}$}
   \STATE{Update $\phi_r,\phi_d$, and $\phi$ from gradients $\nabla_{\phi_r}\gL,\nabla_{\phi_d}\gL$, and $\nabla_{\phi}\gL$}
   \ENDWHILE
\end{algorithmic}
\end{algorithm}

One concern in the superposition of $\gL_{task}$ and $\gL_{rule}$ is their scale differences that may cause all learnable parameters to be dominated by one particular objective regardless of $\alpha$, and hence become unbalanced.
This is not a desired behavior as it significantly limits the expressiveness of {\ours} and may cause convergence into a single mode.
E.g., if $\gL_{rule}$ is much larger than $\gL_{task}$, then {\ours} will become a rule-based model even when $\alpha$ is close to 0.
To minimize such imbalanced behavior, we propose to adjust the scaling automatically.
Before starting a learning process, we compute the initial loss values  $\gL_{rule,0}$ and $\gL_{task,0}$ on a training set and introduce a scale parameter $\rho=\gL_{rule,0}/\gL_{task,0}$. 
Overall, the {\ours} objective function becomes:
\begin{align}
    \gL=E_{\alpha \sim P(\alpha)} [\alpha\gL_{rule} + \rho(1-\alpha)\gL_{task}] \label{eq:scaling}.
\end{align}

{\ours} is model-type agnostic -- we can choose and appropriate inductive bias for encoders and the decision block based on the type and task.
E.g. if the input data is an image and the task is classification, the encoders can be convolutional layers to extract hidden representations associated with local spatial coherence, and the decision block can be an MLP followed by a softmax layer.

\section{Integrating Rules via Input Perturbations}
\label{sec:rule_loss}
\vspace{-0.05in}

Algorithm~\ref{alg} requires a rule-based objective $\gL_{rule}$ that is a function of $(\vx,\hat{\vy})$ and is only differentiable with respect to the learnable parameters of the model. 
In some cases, it is straightforward to convert a rule into a differentiable form. 
For example, for a rule defined as $r(\vx, \hat{\vy}) \leq \tau$ given a differentiable function $r(\cdot)$, we can propose $\gL_{rule} = \text{max}(r(\vx, \hat{\vy}) - \tau, 0)$ that has a penalty with an increasing amount as the violation increases.
However, there are many valuable rules that are non-differentiable with respect to the input $\vx$ or learnable parameters, and in these cases it may not be possible to define a continuous function $\gL_{rule}$ as above.
Some examples include expressive statements represented as concatenations of Boolean rules (e.g. fitted decision trees)~\cite{dash2018boolean}, such as \textit{`The probability of the $j$-th class $\hat{\vy}_j$ is higher when $a<\vx_k$ (where $a$ is a constant and $\vx_k$ is the $k$-th feature)'} or \textit{`The number of sales is increased when price of an item is decreased.'}.

\begin{algorithm}[!htbp]
  \caption{{\ours} via perturbation-based integration of rules.}
   \label{alg:perturbation}
   \begin{flushleft}
    \textbf{Input}: Training data $\gD=\{(\vx_i,\vy_i):i=1,\cdots,N\}$.\\
    \textbf{Output}: Optimized parameters\\
   \textbf{Require}: Rule encoder $\phi_r$, data encoder $\phi_d$, decision block $\phi$, and distribution $P(\alpha)$.\\
   \end{flushleft}
   \begin{algorithmic}[1]
   \STATE Initialize $\phi_r,\phi_d$, and $\phi$
   \WHILE{not converged}
   \STATE{Get mini-batch $\gD_b$ from $\gD$ and $\alpha_b\in\sR$ from $P(\alpha)$}
   \STATE Get perturbed input $\vx_p=\vx+\delta\vx$ where $\vx\in\gD_b$
   \STATE Get $\vy$ and $\vy_p$ from $\vx$ and $\vx_p$ through $\phi_r,\phi_d,\phi,$ and $\alpha_b$, respectively
   \STATE Define $\gL_{rule}=\gL_{rule}(\vx,\vx_p,\hat{\vy},\hat{\vy}_p)$ based on a rule and $\gL_{task}=\gL_{task}(\vy,\hat{\vy})$ to compute\\ $\gL=\alpha_b\gL_{rule} + (1-\alpha_b)\gL_{task}$
   \STATE{Update $\phi_r,\phi_d$, and $\phi$ from gradient $\nabla_{\phi_r}\gL,\nabla_{\phi_d}\gL$, and $\nabla_{\phi}\gL$}
   \ENDWHILE
\end{algorithmic}
\end{algorithm}

\begin{figure*}[!htbp]
    \centering
    \begin{subfigure}[b]{.32\textwidth}
        \centering
        \includegraphics[width=0.95\textwidth]{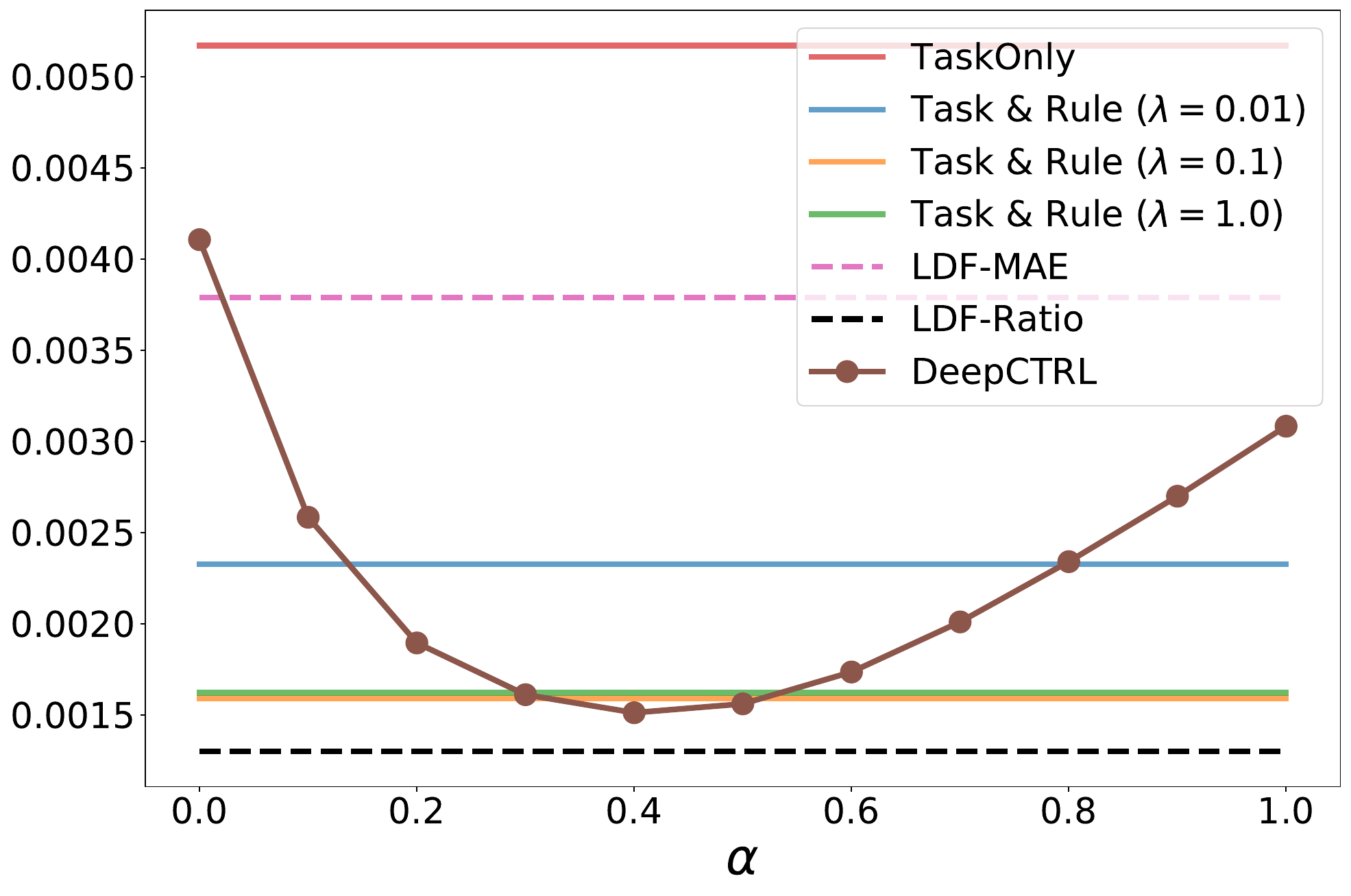}
        \caption{MAE}\label{fig:dp-mae}
    \end{subfigure}
    \hfill
    \begin{subfigure}[b]{.32\textwidth}
        \centering
        \includegraphics[width=0.915\textwidth]{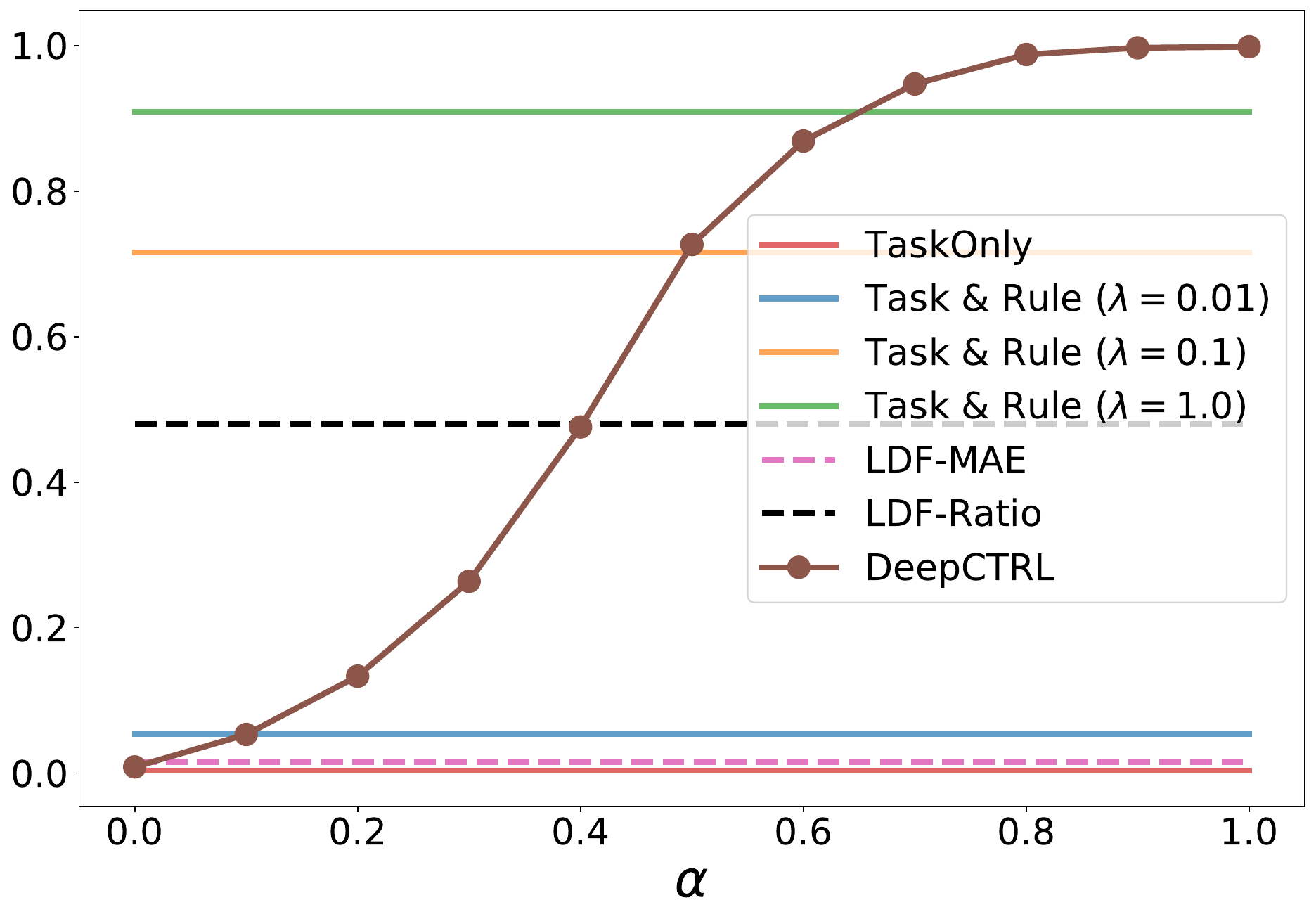}
        \caption{Verification ratio}\label{fig:dp-verification-ratio}
    \end{subfigure}
    \hfill
    \begin{subfigure}[b]{.32\textwidth}
        \centering
        \includegraphics[width=0.95\textwidth]{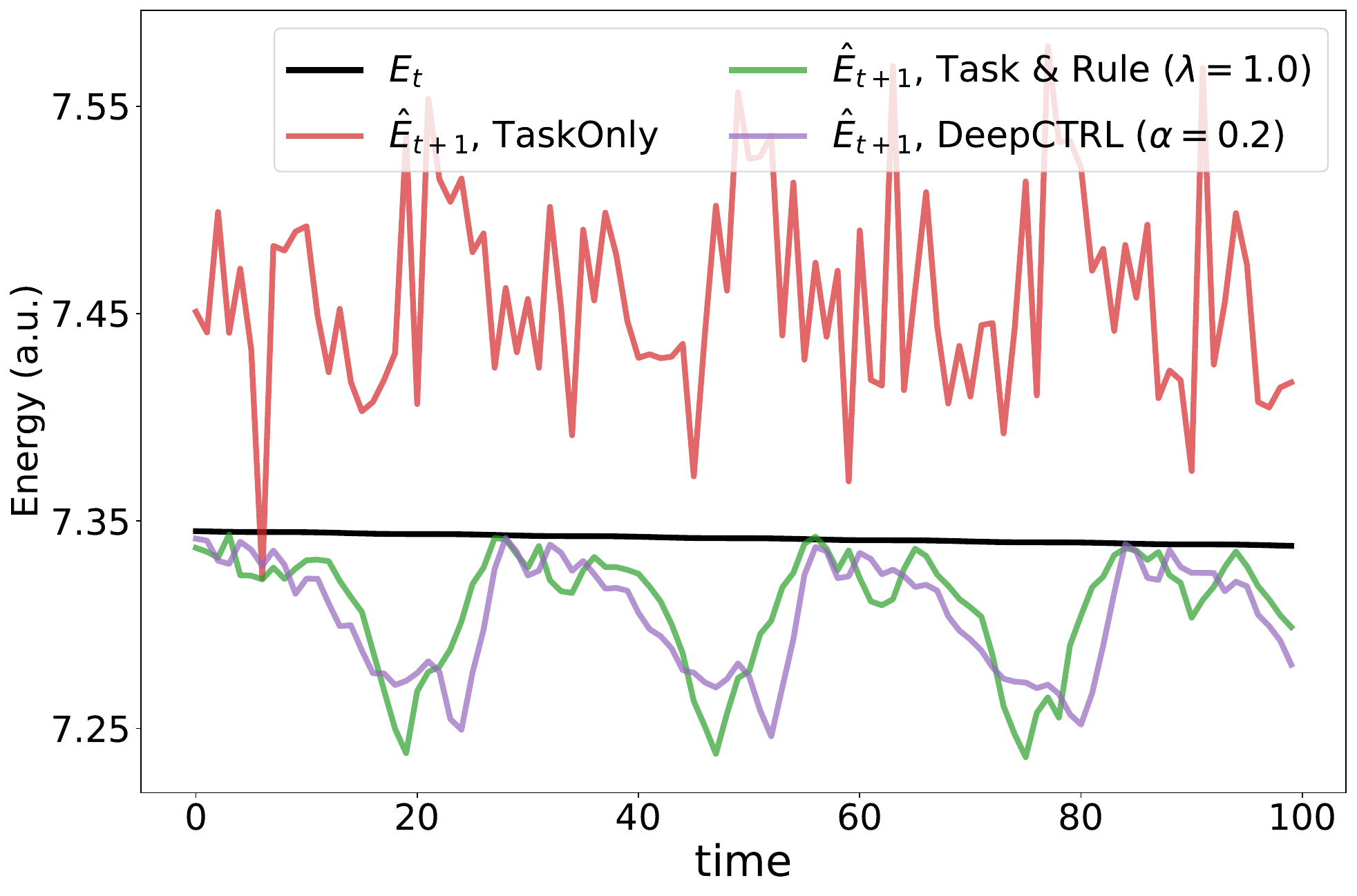}
        \caption{Energy prediction}\label{fig:dp-energy}
    \end{subfigure}
    \caption{Experimental results for the double pendulum task. (Left) Task-based prediction error and (Middle) verification ratio from different models. {\ours} has a scale parameter which adjusts the scale mismatch between $\gL_{task}$ and $\gL_{rule}$. The performances of \textsc{LDF} method are highly sensitive to hyperparameter and its output is not reliable. (Right) Current and predicted energy at time $t$ and $t+1$, respectively. According to the energy damping rule, $\hat{E}_{t+1}$ should not be larger than $E_t$.}\label{fig:dp-result}
\end{figure*}

We introduce an input perturbation method, overviewed in Algorithm~\ref{alg:perturbation}, to generalize {\ours} to non-differentiable constraints.
The method is based on introducing a small perturbation $\delta\vx$ (Line 4 in Algorithm 2) to input features $\vx$ in order to modify the original output $\hat{\vy}$ and construct the rule-based constraint $\gL_{rule}$ for it.
For example, if we were to incorporate the first sample rule in the previous paragraph (concatenations of Boolean rules), we would only consider $\vx_p$ as a valid perturbed input when $\vx_k<a$ and $a<\vx_{p,k}$ and $\hat{\vy}_p$ is computed from the $\vx_p$.
$\gL_{rule}$ is defined as:
\begin{equation}
    \gL_{rule}(\vx,\vx_p,\hat{\vy}_j,\hat{\vy}_{p,j}) = \text{ReLU}(\hat{\vy}_j - \hat{\vy}_{p,j})\cdot I(\vx_k<a)\cdot I(\vx_{p,k}>a).
\end{equation}
This perturbation-based loss function is combined with a task-based loss function (Line 6 in Algorithm~\ref{alg:perturbation}) to update all parameters in $\phi_r,\phi_d$, and $\phi$.
The control parameter $\alpha$ is used for both $\hat{\vy}$ and $\hat{\vy}_p$, and $\gL_{task}$ only considers the original output $\hat{\vy}$.
All learnable parameters in the rule encoder, data encoder, and decision block are shared across the original input $\vx$ and the perturbed input $\vx_p$. 
Overall, this perturbation-based method expands the scope of rules we can incorporate into {\ours}.


\section{Experimental Results}
\label{sec:exp}
\vspace{-0.05in}

We evaluate {\ours} on machine learning use cases from Physics, Retail, and Healthcare, where utilization of rules is particularly important. 

For the rule encoder ($\phi_r$), data encoder ($\phi_d$), and decision block ($\phi$), we use MLPs with ReLU activation at intermediate layers, similarly to \cite{breen2019newton,greydanus2019hamiltonian}. 
We compare {\ours} to the {\dataonly} baseline, which is trained with fixed $\alpha=0$, i.e. it only uses a data encoder ($\phi_d$) and a decision block ($\phi$) to predict a next state.
In addition, we include {\dataonly} with rule regularization, {\taskwithrule}, based on Eq.~\ref{eq:convention} with a fixed $\lambda$.
We make a comparison to Lagrangian Dual Framework (LDF)~\cite{fioretto2020lagrangian} that enforces rules by solving a constraint optimization problem.

\subsection{Improved Reliability Given Known Principles}\label{sec:dp}
\vspace{-0.05in}
By adjusting the control parameter $\alpha$, a higher rule verification ratio, and thus more reliable predictions, can be achieved. 
Operating at a better verification ratio could be beneficial for performance, especially if the rules are known to be (almost) always valid as in natural sciences. 
We demonstrate a systematic framework to encode such principles from known literature, although they may not be completely learnable from training data.

\textbf{Dataset, task and the rule:}
Following ~\cite{asseman2018learning}, we consider the time-series data generated from double pendulum dynamics with friction, from a given initial state $(\theta_1,\omega_1,\theta_2,\omega_2)$ where $\theta,\omega$ are angular displacement and velocity, respectively.
We first generate the time-series data with a high sampling frequency (200Hz) to avoid numerical errors, and then downsample to a lower sampling frequency (10Hz) to construct the training dataset.
We define the task as predicting the next state $\vx_{t+1}$ of the double pendulum from the current state $\vx_t=(\theta_{1t},\omega_{1t},\theta_{2t},\omega_{2t})$.
We construct a synthetic training dataset using the analytically-defined relationship between inputs and outputs, and introduce a small additive noise ($\epsilon\sim\gN(0,0.0001)$) to model measurement noise in the setup.
We focus on teaching the rule of energy conservation law.
Since the system has friction, $E(\vx_t)>E(\vx_{t+1})$, where $E(\vx)$ is the energy of a given state $\vx$.
We apply an additional constraint based on the law such that $\gL_{rule}(\vx,\hat{\vy})=\text{ReLU}(E(\hat{\vy})-E(\vx))$.
To quantify how much the rule is learned, we evaluate the `verification ratio', defined as the ratio of samples that satisfy the rule. 

\textbf{Results on accuracy and rule teaching efficacy:}
Fig.~\ref{fig:dp-result} shows how the control parameter affects the task-based metric and the rule-based metric, respectively.
The additionally-incorporated rule in {\ours} is hugely beneficial for obtaining more accurate predictions of the next state, $\hat{\vx}_{t+1}$.
Compared to {\dataonly}, {\ours} reduces the prediction MAE significantly -- the parameters are driven to learn better representations with the domain knowledge constraints.
Moreover, {\ours} provides much higher verification ratio than {\dataonly} and the resultant energy is not only verified, but also more stable (close to that of {\taskwithrule}) (Fig.~\ref{fig:dp-energy}).
It demonstrates that {\ours} is able to incorporate the domain knowledge into a data-driven model, providing reliable and robust predictions.
{\ours} allows to control model's behavior by adjusting $\alpha$ without retraining: when $\alpha$ is close to 0 in Fig.~\ref{fig:dp-mae} and~\ref{fig:dp-verification-ratio}, {\ours} is in the \textit{task only} region where its rule-based metric is degraded; by increasing $\alpha$, the model's behavior is more dominated by the rule-based embedding $\vz_r$ -- i.e. although the prediction error is increased, the output is more likely to follow the rule.
However, {\ours} is still better than {\dataonly} in both metrics regardless of the value of $\alpha$.

\textbf{Comparison to baselines:}
Fig.~\ref{fig:dp-result}(a, b) shows a comparison of {\ours} to the baselines of training with a rule-based constraint as a fixed regularization term.
We test different $\lambda\in\{0.01,0.1,1.0\}$ and all show that the additive regularization (Eq.~\ref{eq:convention}) is helpful for both aspects.
The highest $\lambda$ provides the highest verification ratio, however, the prediction error is slightly worse than that of $\lambda=0.1$.
We find that the lowest prediction error of the fixed baseline is comparable to (or even larger than) that of {\ours}, but the highest verification ratio of the fixed baseline is still lower than that of {\ours}.
The computed energy values from the fixed baseline are also similar to those from {\ours}.
In addition, we consider the benchmark of imposing the rule-constraint with LDF and demonstrate two results where its hyperparameters are chosen by the lowest MAE (\textsc{LDF-MAE}) and highest rule verification ratio (\textsc{LDF-Ratio}) on validation set, respectively. We note that LDF does not allow the capability of flexibly changing the rule strength at inference as {\ours}, so it needs to be retrained for different hyperparameters to find such operation points.
\textsc{LDF-MAE} yields higher MAE and lower rule verification ratio compared to others on the test set, showing lack of generalizability of the learned rule behavior.
On the other hand, \textsc{LDF-Ratio} provides lower MAE than others. However, only 50\% of outputs follow the rule which significantly lowers the reliability of the method.
Overall, these results demonstrate that {\ours} is competitive in MAE compared to fixed methods like LDF, while providing the flexibility of adjusting the rule strength to operate at a more favorable point in terms of rule verification ratio, and enabling the extra capabilities presented next.


\textbf{Scale of the objectives:}
It is important that the scales of $\gL_{rule}$ and $\gL_{task}$ are balanced in order to avoid all learnable parameters becoming dominated by one of the objective.
With the proposed scale parameter $\rho$, the combined loss $\gL$ has to be of a similar scale for the two extremes ($\alpha\rightarrow0$ and $\alpha\rightarrow1$). 
Fig.~\ref{fig:dp-mae} shows that this scaling enables a more U-shape curve, and the verification ratio when $\alpha\rightarrow0$ is close to that of {\dataonly} in Fig.~\ref{fig:dp-verification-ratio}.
In other words, {\ours} is close to {\taskonly} as $\alpha\rightarrow0$ and close to {\ruleonly} as $\alpha\rightarrow1.0$, which is a model trained with a fixed $\alpha=1.0$.
It is not necessary to search the scale parameter before training; instead, it is fixed at the beginning of training with the initial loss values of $\gL_{rule}$ and $\gL_{task}$.

\textbf{Optimizing rule strength on validation set:}
Users can run inference with {\ours} with any value for $\alpha$ without retraining. 
In Fig.~\ref{fig:dp-result} we consider the scenario where users pick an optimal $\alpha$ based on a target verification ratio. 
For a goal of $>90\%$ verification ratio on the validation set, $\alpha>0.6$ is needed, and this guarantees a verification ratio of $80.2\%$ on the test set. 
On the other hand, if we consider optimization of $\lambda$ of the fixed objective function (`Task \& Rule') benchmark for $>90\%$ validation verification ratio, we observe a test verification ratio of $71.6\%$, which is $8.6 \%$ lower than {\ours}. 
In addition, the verification ratio on the validation set can be used to determine the rule strength that minimizes error. 
On the validation set, minimum MAE occurs when $\alpha=0.2$, which corresponds to a rule verification ratio around 60 \%. 
If we search for the $\alpha$ that satisfies the same rule verification ratio of 60 \% on the test set, we observe that $\alpha=0.4$ is optimal, which also gives the lowest MAE. 
In other words, the rule verification ratio is a robust unsupervised model selection proxy for {\ours}, underlining the generalizability of the learned representations.

\subsection{Examining Candidate Rules} \label{sec:m5}
\vspace{-0.05in}
{\ours} allows `hypothesis testing' for rules. 
In many applications, rules are not scientific principles but rather come from insights, as in economical or sociological sciences.
We do not always want rules to dominate data-driven learning as they may hurt accuracy when they are not valid.

\textbf{Dataset, task and the rule:}
We focus on the task of sales forecasting of retail goods on the M5 dataset~\footnote{\url{https://www.kaggle.com/c/m5-forecasting-accuracy/}}.
While the original task is forecasting daily sales across different goods at every store, we change the task to be forecasting weekly sales since the prices of items are updated weekly. 
For this task, we consider the economics principle as the rule \cite{Browning1986MicroeconomicTA}: \textit{price-difference and sales-difference should have a negative correlation coefficient}: $r=\frac{\Delta\textsc{sales}}{\Delta\textsc{prices}}<0.0$.
This inductive bias is incorporated using the perturbation-based objective described in Sec.~\ref{sec:rule_loss}.
The positive perturbation is applied to \textit{price} of an input item, and $\gL_{rule}$ is a function of the perturbed output $\hat{\vy}_p$ and the original output $\hat{\vy}$: $\gL_{rule}(\hat{\vy},\hat{\vy}_p)=\text{ReLU}(\hat{\vy}_p-\hat{\vy})$.
Unlike the previous task, the rule of this task is \textit{soft} -- not all items are known to follow the rule and their correlation coefficients are different.

\textbf{Experimental setting:}
We split items in three groups: (1) {\negone} items with a negative price-sales correlation ($r<-0.1$), (2) {\negtwo} items with negative price-sales correlation ($r<-0.2$), and (3) {\negthree} items with negative correlation ($r<-0.3$).
We train {\dataonly}, LDF, and {\ours} on each group to examine how beneficial the incorporated rule is.
Additionally, we examine candidate models by applying a model trained on {\negtwo} to unseen groups: (1) {\negone}, (2) {\postwo}, and (3) {\posthree} where the rule is weak in the latter two groups. 


\begin{figure*}[t]
    \centering
    \begin{subfigure}[b]{.3\linewidth}
        \centering
        \includegraphics[width=\textwidth]{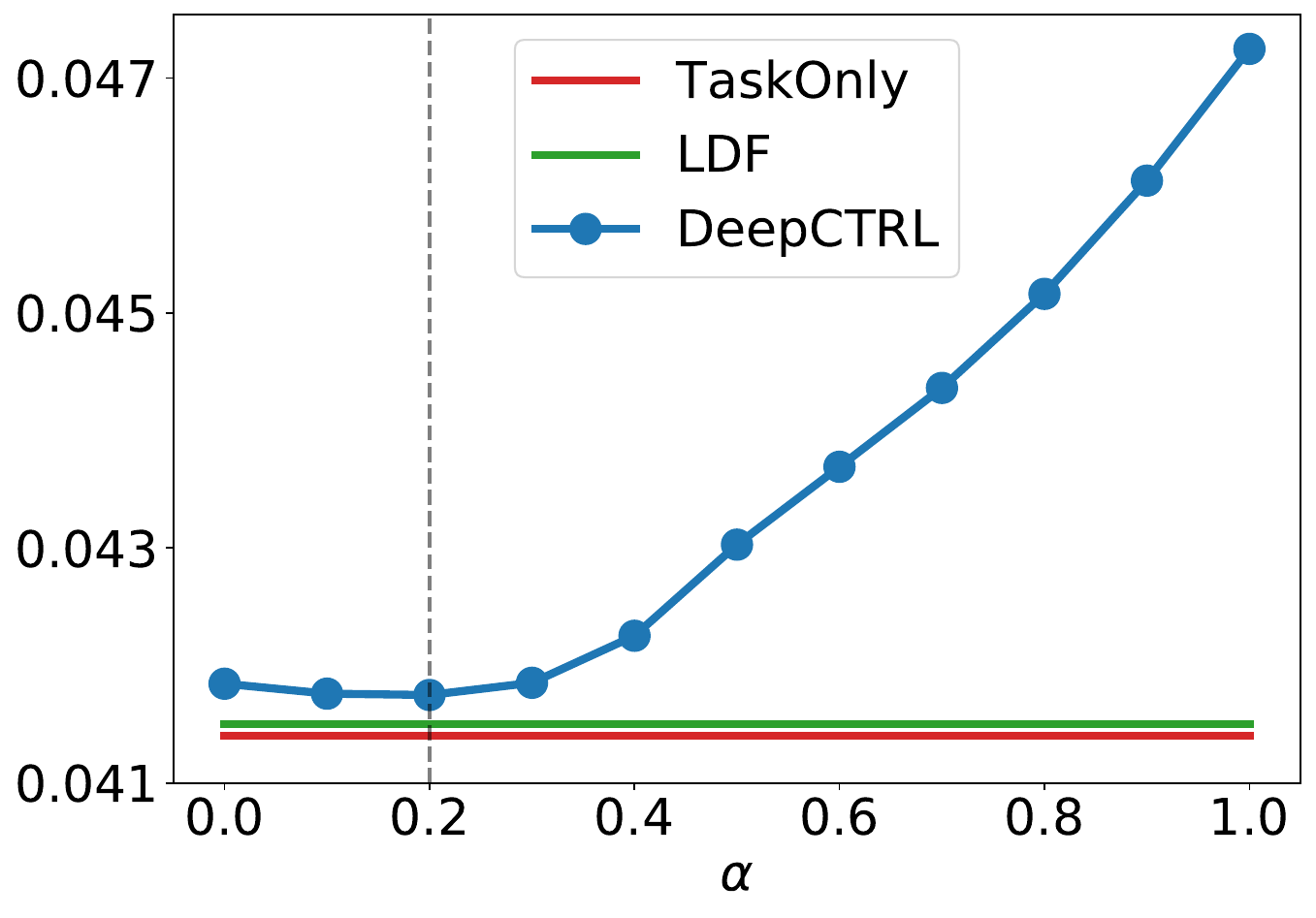}
        \caption{Train/test on {\negone}}
        \label{fig:m5-mae-0.1}
    \end{subfigure}
    \hfill
    \begin{subfigure}[b]{.3\linewidth}
        \centering
        \includegraphics[width=\textwidth]{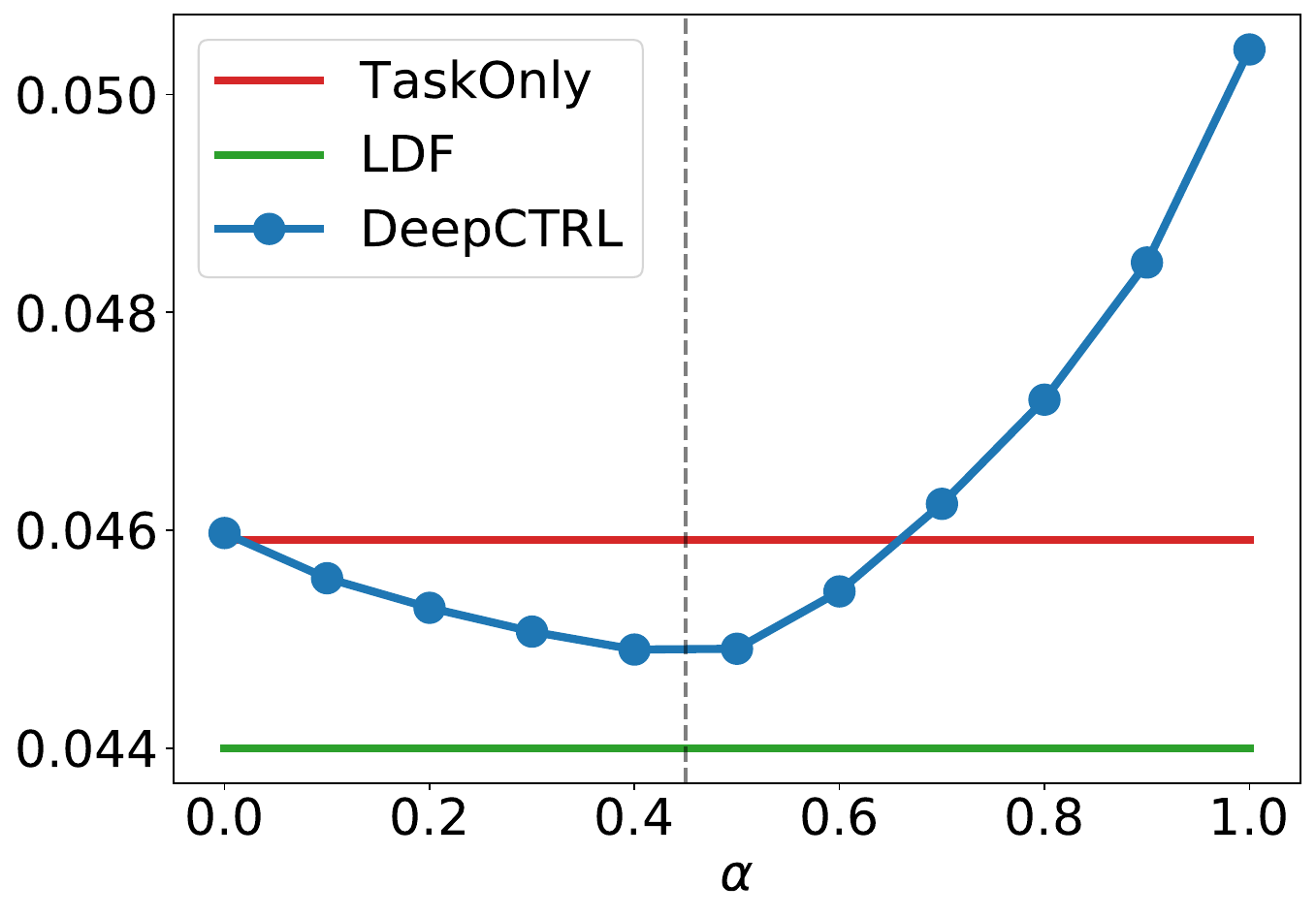}
        \caption{Train/test on \negtwo} \label{fig:m5-mae-0.2}
    \end{subfigure}
    \hfill
    \begin{subfigure}[b]{.3\linewidth}
        \centering
        \includegraphics[width=\textwidth]{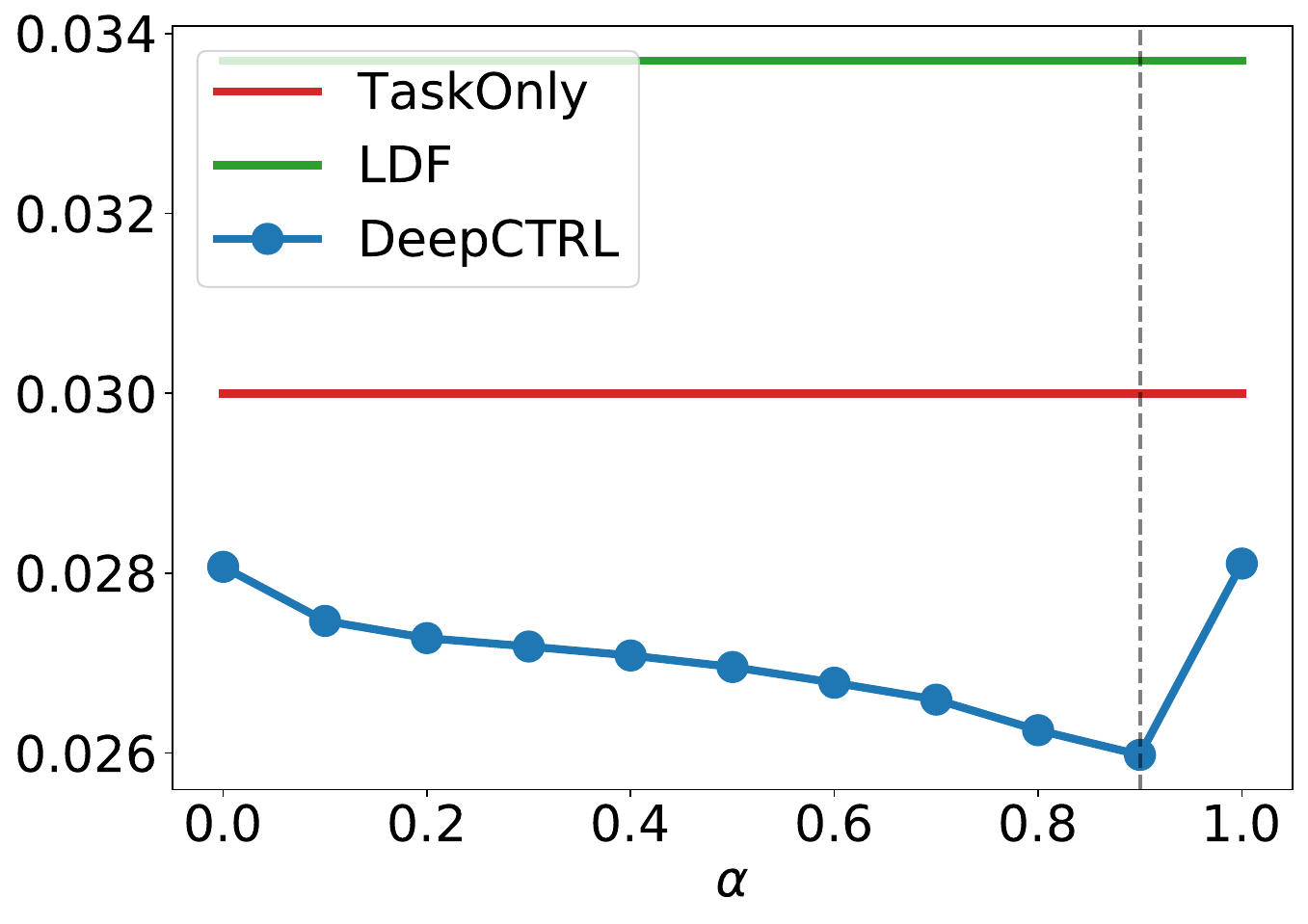}
        \caption{Train/test on \negthree}\label{fig:m5-mae-0.3}
    \end{subfigure}
    \caption{Candidate rule testing. Sales prediction error (MAE) across three groups with different correlation coefficients between $\Delta\textsc{sales}$ and $\Delta\textsc{prices}$. The dashed lines point the optimal $\alpha$ providing the lowest error.} \label{fig:m5-exam-rule}
\end{figure*}

\begin{figure*}[t]
    \centering
    \begin{subfigure}[b]{.3\linewidth}
        \centering
        \includegraphics[width=\textwidth]{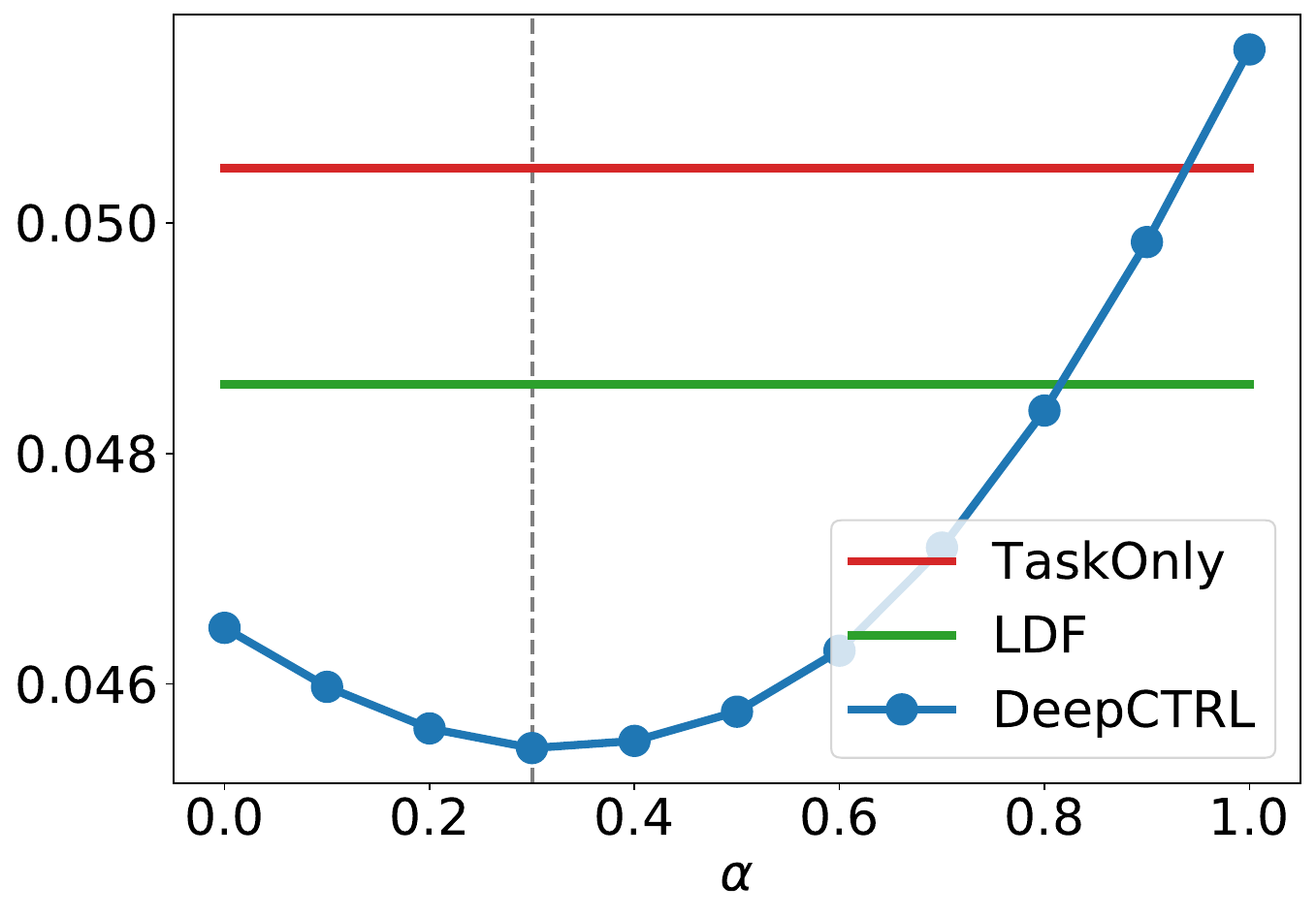}
        \caption{Train: {\negtwo}, Test: {\negone}} \label{fig:m5-mae_-0.2_-0.1}
    \end{subfigure}
    \hfill
    \begin{subfigure}[b]{.3\linewidth}
        \centering
        \includegraphics[width=\textwidth]{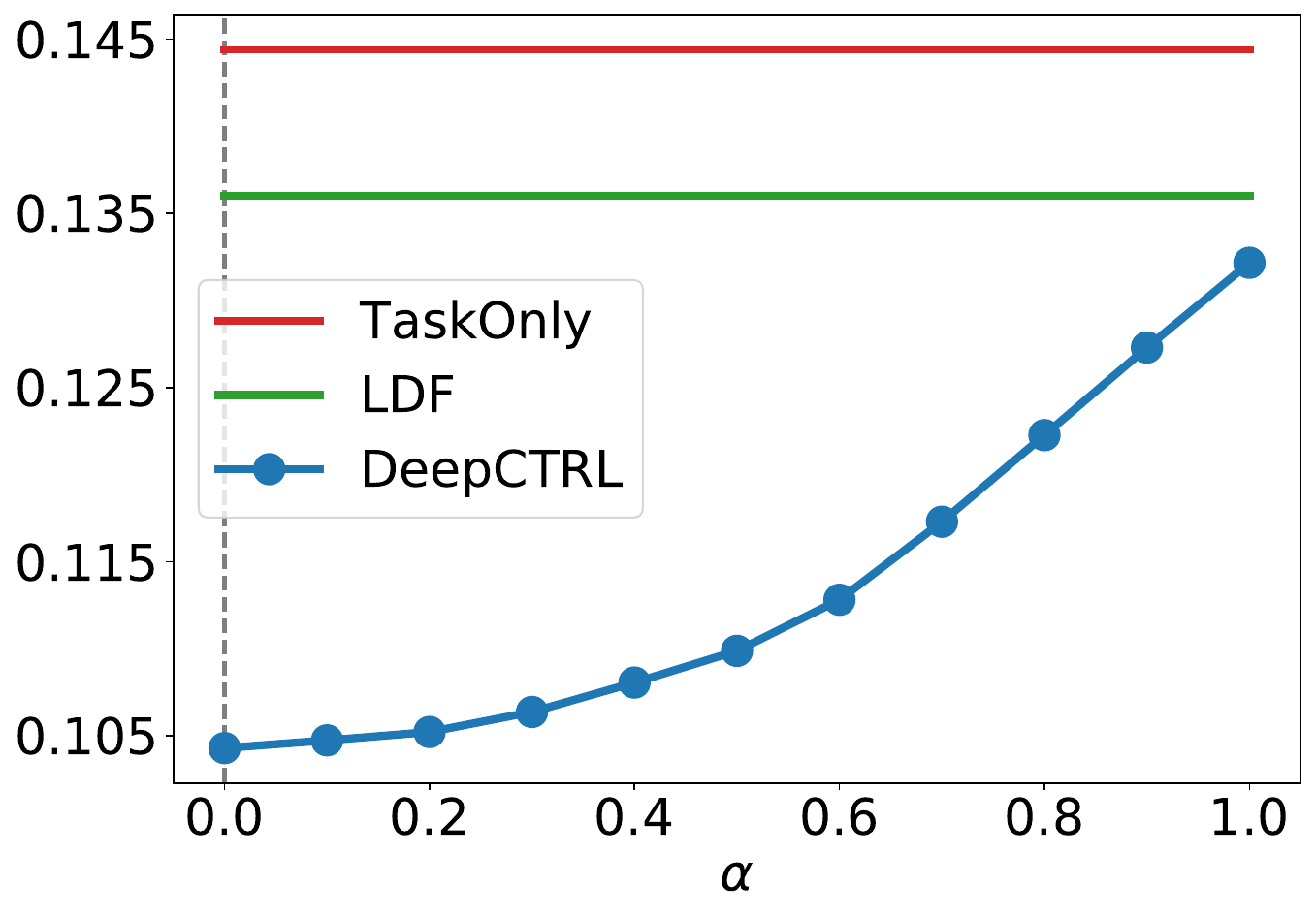}
        \caption{Train: {\negtwo}, Test: {\postwo}}  \label{fig:m5-mae_-0.2_0.2}
    \end{subfigure}
    \hfill
    \begin{subfigure}[b]{.3\linewidth}
        \centering
        \includegraphics[width=\textwidth]{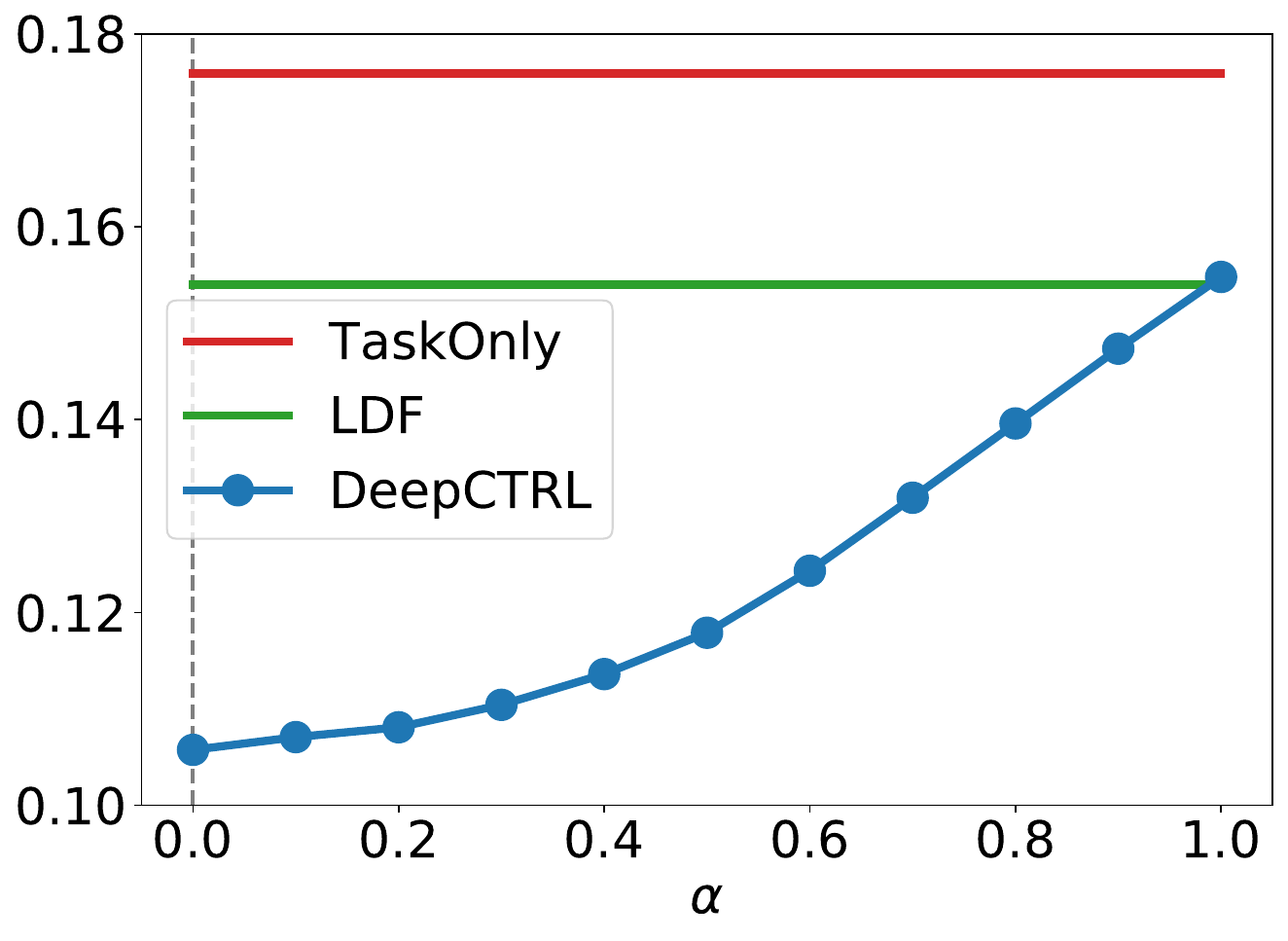}
        \caption{Train: {\negtwo}, Test: {\posthree}} \label{fig:m5-mae_-0.2_0.3}
    \end{subfigure}
    \caption{Candidate model testing. Sales prediction error (MAE) for three groups from a model trained on {\negtwo}. Note that the correlation coefficients in the target groups are different to that of the source group. The dashed lines point the optimal $\alpha$ providing the lowest error.} \label{fig:m5-exam-model}
\end{figure*}

\textbf{Candidate rule testing:} 
Fig.~\ref{fig:m5-exam-rule} shows the weekly sales prediction results.
Compared to {\dataonly}, {\ours} obtains lower MAE on {\negtwo} and {\negthree}, but {\dataonly} outperforms {\ours} when the items have weaker correlations between $\Delta\textsc{sales}$ and $\Delta\textsc{prices}$.
This is reasonable since the rule we impose is more dominant in {\negtwo} and {\negthree}, however, it is less informative for items in {\negone}.
While LDF provides lower MAE on {\negtwo}, its performance is significantly degraded on {\negthree} due to an unstable dual learning process.
Furthermore, the rule-dependency can be discovered via the value of $\alpha$ providing the lowest error.
Fig.~\ref{fig:m5-exam-rule} clearly demonstrates that the optimal $\alpha$ is shifted to larger values as the correlation gets stronger.
In other words, as items have stronger correlations between $\Delta\textsc{prices}$ and $\Delta\textsc{sales}$, the corresponding rule is more beneficial, and thus the rule-dependency is higher.
The post-hoc controllability of {\ours} enables users to examine how much the candidate rules are informative for the data.

\textbf{Candidate model testing:} 
The hypothesis testing use case can be extended to model testing to evaluate appropriateness of a model for the target distribution.
Fig.~\ref{fig:m5-exam-model} shows models testing capability, where items in each domain have different price-sales correlations.
Overall, {\ours} is superior to other methods like LDF as it learns the task and rule representations in disentangled ways, so that the negative impact of a rule is more minimal via a lower when the rule is not helping for the task.
There are notable points from the experimental results.
First, Fig.~\ref{fig:m5-mae_-0.2_-0.1} shows that {\dataonly}, LDF, and {\ours} can be applicable to target domain (\negone) without significant performance degradation.
Compared to Fig.~\ref{fig:m5-mae-0.1}, MAE from {\dataonly} is increased from 0.041 to 0.05 because {\dataonly} is optimized on {\negtwo}.
While {\ours} is also degraded (0.042 to 0.045), the gap is much smaller than that of {\dataonly} as we can optimize {\ours} by tuning $\alpha$.
Second, compared to Fig.~\ref{fig:m5-mae-0.2}, we see that the optimal $\alpha$ is decreased to 0.3 from 0.45, showing that rule-dependency becomes weaker compared to source domain.
Third, if the model is applied to ({\postwo} and {\posthree}) where the imposed rule for the source domain is not informative at all, not only the prediction quality is significantly degraded, but also the rule-dependency is minimized (the optimal $\alpha$ is 0).
This indicates the benefit of controllability to examine how much the existing rules or trained models are appropriate for given test samples.

\subsection{Adapting to Distribution Shifts using the Rule Strength} \label{sec:da}
\vspace{-0.05in}

There may be shared rules across multiple datasets for the same task -- e.g. the energy conservation law must be valid for pendulums of different lengths or counts. 
On the other hand, some rules may be valid with different strengths among different subsets of the data -- e.g. in disease prediction, the likelihood of cardiovascular disease with higher blood pressure increases more for older patients than younger patients. 
When the task is shared but data distribution and the validity of the rule differ among different datasets, {\ours} is useful for adapting to such distribution shifts by controlling $\alpha$, avoiding then need for fine-tuning or training from scratch. 

\textbf{Dataset, task and the rule:}
We evaluate this aspect with a cardiovascular classification task\footnote{\url{https://www.kaggle.com/sulianova/cardiovascular-disease-dataset}}.
We consider 70,000 records of patients with half having a cardiovascular disease.
The target task is to predict whether the cardiovascular disease is present or not based on 11 continuous and categorical features.
Given that higher systolic blood pressure is known to be strongly associated with the cardiovascular disease \cite{KANNEL2000251}, we consider the rule ``\textit{$\hat{y}_{p,i} > \hat{y}_i$ if $x_{p,i}^{press} > x_i^{press}$} ", where $x_i^{press}$ is systolic blood pressure of the $i$-th patient, and $\hat{y}_i$ is the probability of having the disease. The subscript $p$ denotes perturbations to the input and output.
Fitting a decision tree, we discover that if blood pressure is higher than 129.5, more than 70\% patients have the disease. 
Based on this information, we split the patients into two groups: (1) $\{i\ :\ \{x_i^{press}<129.5\cap y_i=1\}\cup\{x_i^{press}\geq129.5\cap y_i=0\}\}$ (called \textsc{Unusual}) and (2) $\{i\ :\ \{x_i^{press}<129.5\cap y_i=0\}\cup\{x_i^{press}\geq129.5\cap y_i=1\}\}$ (called \textsc{Usual}).
The majority of patients in the \textsc{Unusual} group likely to have the disease even though their blood pressure is relatively lower, and vice versa.
In other words, the rule is not very helpful since it imposes a higher risk for higher blood pressure.
To evaluate the performance, we create different target datasets by mixing the patients from the two groups with different ratios (See Appendix).

\begin{figure}[t]
    \centering
    \begin{subfigure}[b]{.43\linewidth}
        \centering
        \includegraphics[width=\textwidth]{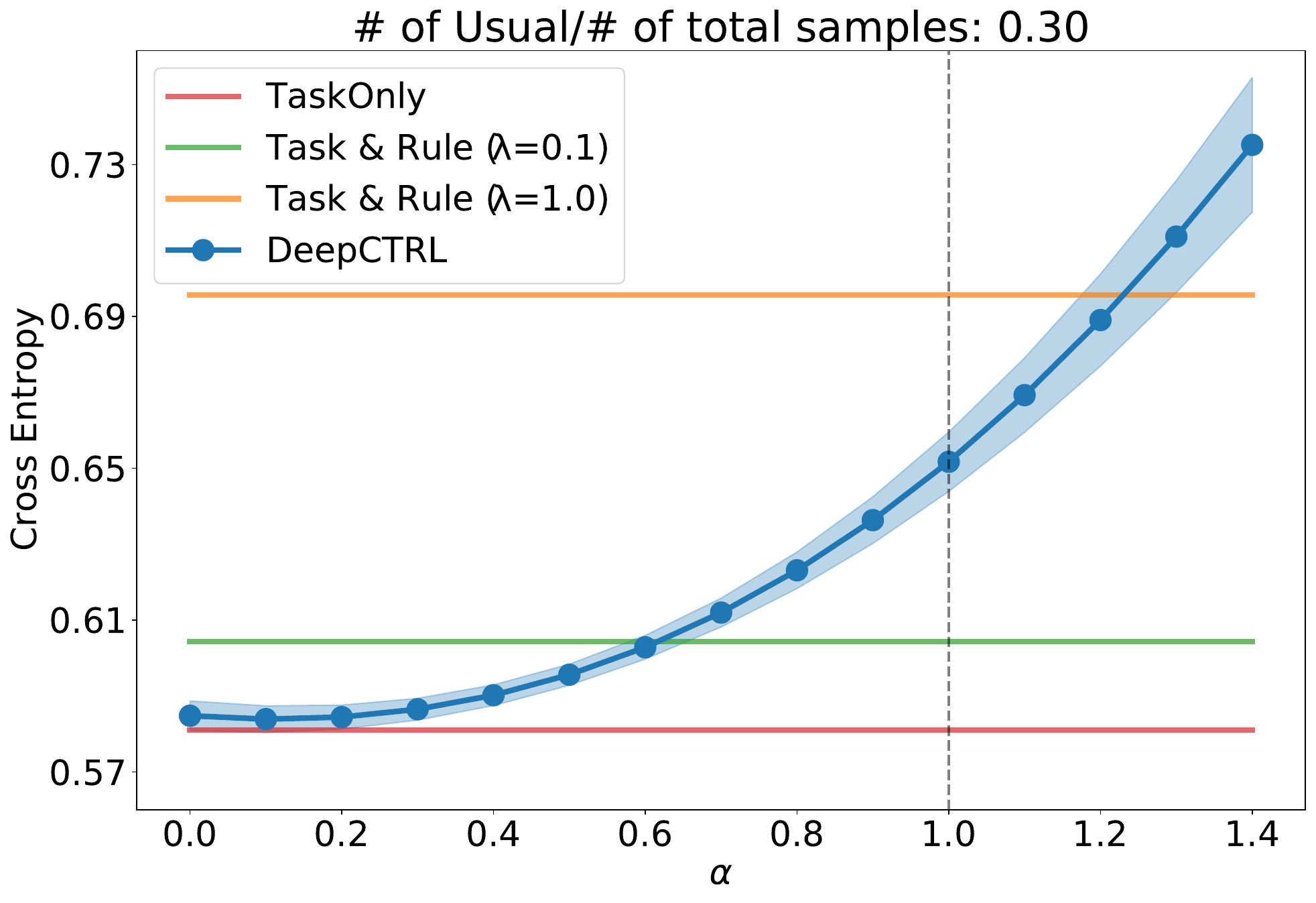}
        \caption{\textsc{Source}}\label{fig:cardio-da-src}
    \end{subfigure}
    \begin{subfigure}[b]{.43\linewidth}
        \centering
        \includegraphics[width=\textwidth]{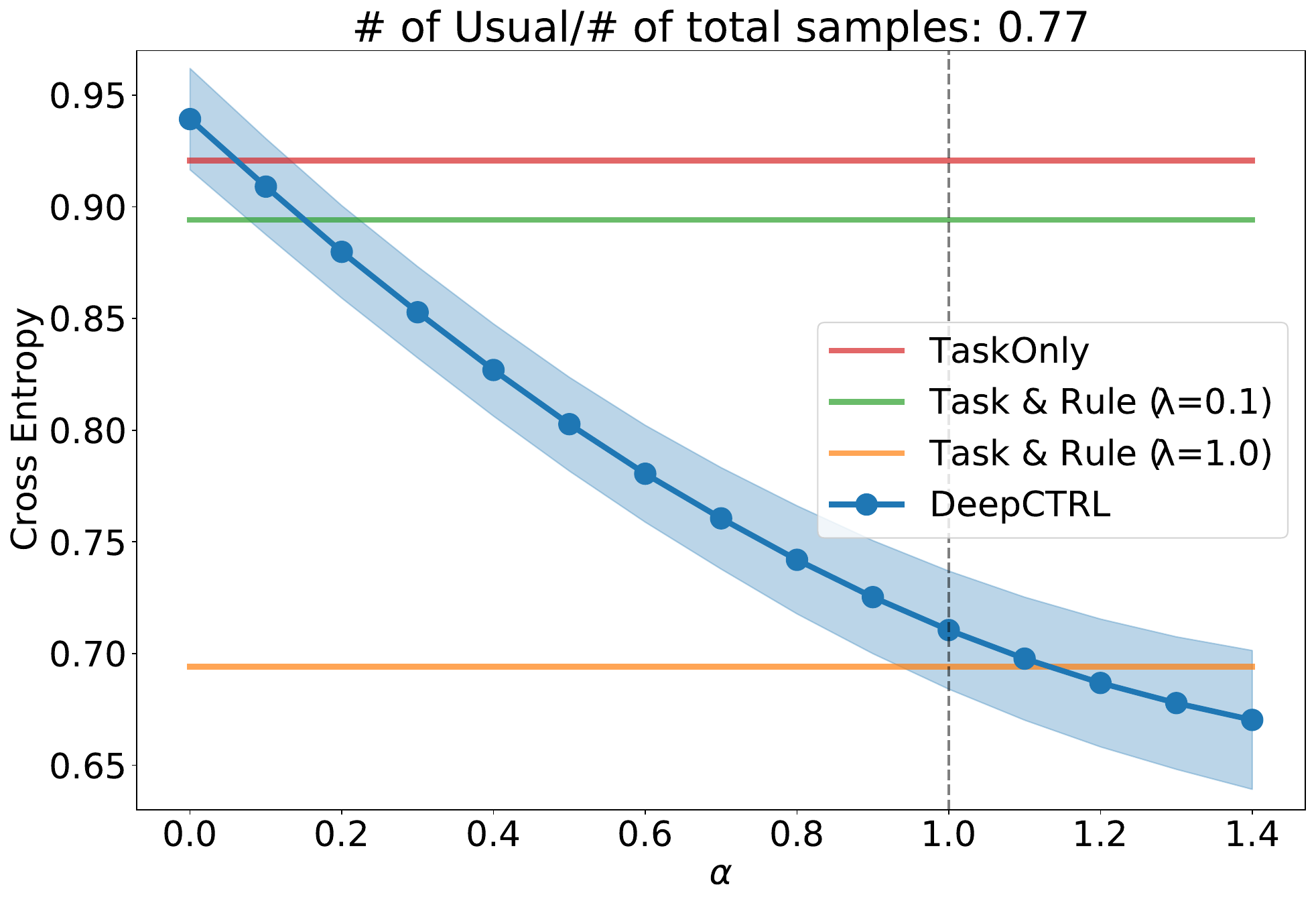}
        \caption{\textsc{Target 1}}\label{fig:cardio-da-target1}
    \end{subfigure}
    \begin{subfigure}[b]{.43\linewidth}
        \centering
        \includegraphics[width=\textwidth]{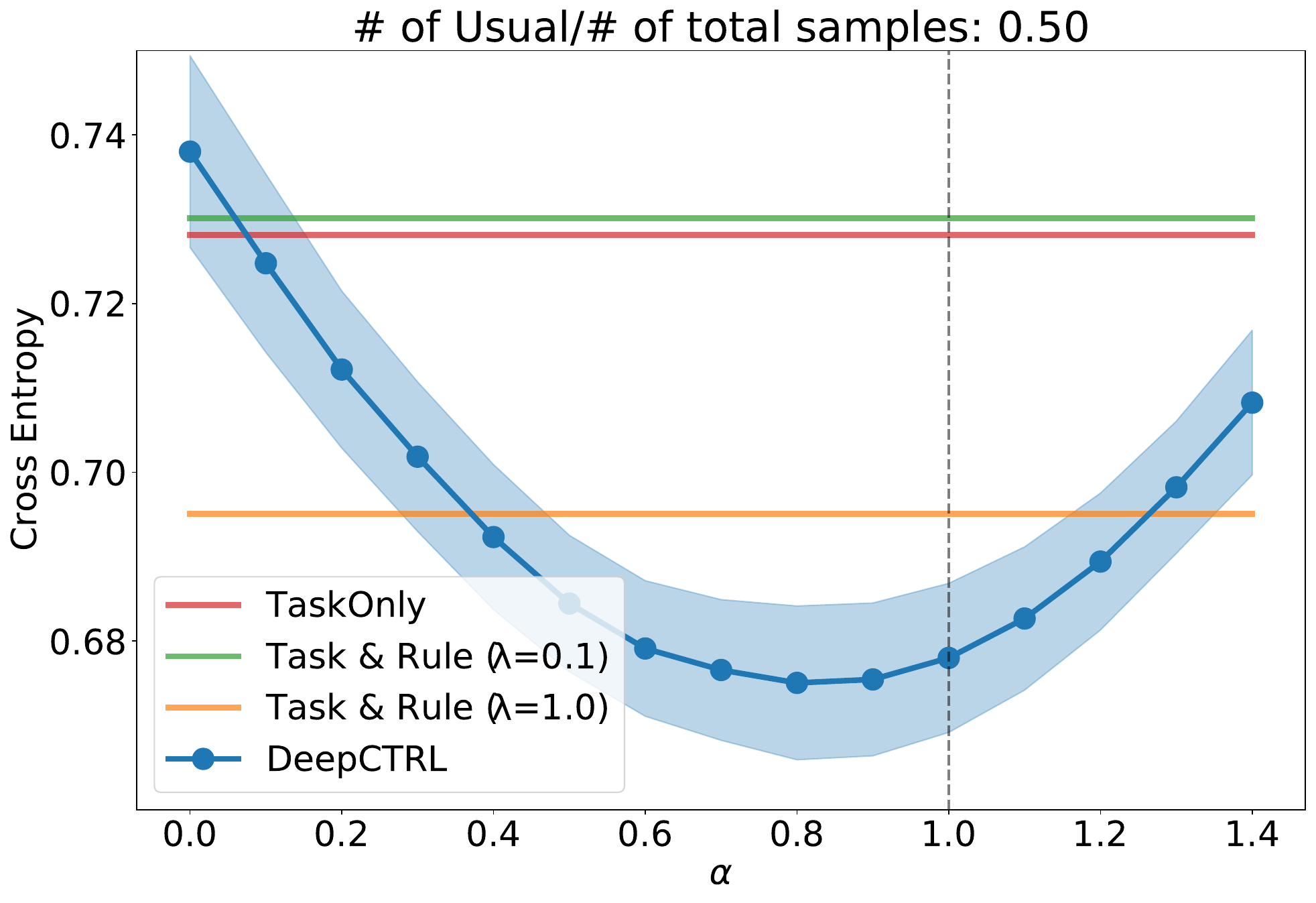}
        \caption{\textsc{Target 2}}\label{fig:cardio-da-target2}
    \end{subfigure}
    \begin{subfigure}[b]{.43\linewidth}
        \centering
        \includegraphics[width=\textwidth]{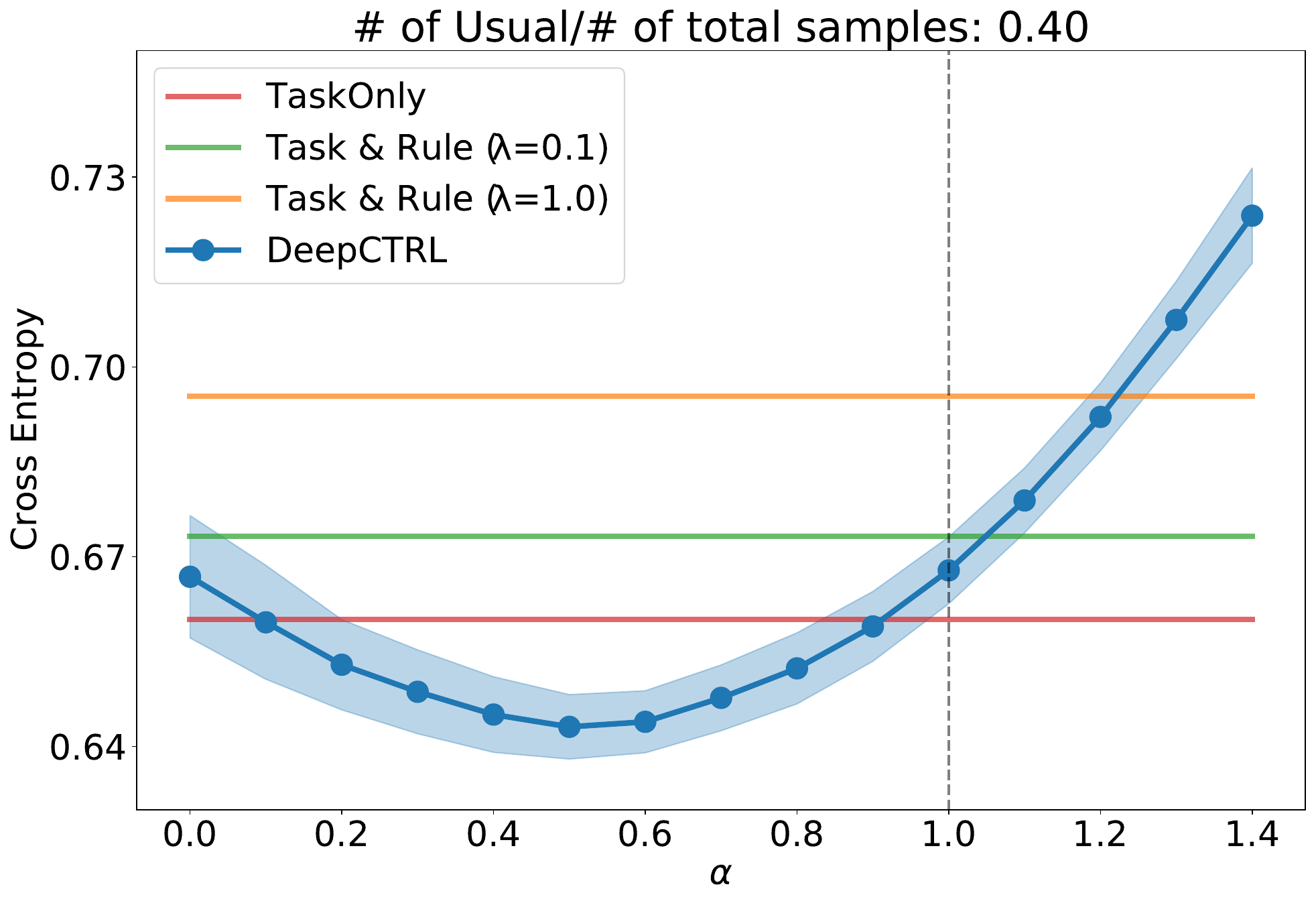}
        \caption{\textsc{Target 3}}\label{fig:cardio-da-target3}
    \end{subfigure}
    \caption{Test cross entropy vs. rule strength for target datasets with varying \textsc{Usual}/\textsc{Unusual} ratio. } \label{fig:cardio-da}
\end{figure}

\textbf{Target distribution performance:}
We first train a model on the dataset from the source domain and then apply the trained model to a dataset from the target domain without any additional training.
Fig.~\ref{fig:cardio-da-src} shows that {\dataonly} outperforms {\ours} regardless of the value of $\alpha$.
This is expected as the source data do not always follow the rule and thus, incorporating the rule is not helpful or even harmful.
Note that the error monotonically increases as $\alpha\rightarrow1$ as the unnecessary (and sometimes harmful) inductive bias gets involved more.
When a trained model is transferred to the target domain, the error is higher due to the difference in data distributions.
However, we show that the error can be reduced by controlling $\alpha$. For \textsc{Target 1} where the majority of patients are from \textsc{Usual}, as $\alpha$ is increased, the rule-based representation, which is helpful now, has more weight and the resultant error decreases monotonically.
When the ratio of patients from \textsc{Usual} is decreased, the optimal $\alpha$ is an intermediate value between 0 and 1.
These demonstrate the capability of {\ours} to adapt the trained model via $\alpha$ towards an optimal behavior if the rule is beneficial for the target domain. Moreover, we can reversely interpret the amount of rule dependency for a target dataset.
As the optimal $\alpha$ is close to 1.0 for \textsc{Target 1}, we expect that the rule is valid for most of the patients, unlike \textsc{Target 3} which has the optimal $\alpha$ around 0.5. 

\textbf{Extrapolating the rule strength:}
While $\alpha$ is sampled between 0 and 1 during training, the encoder outputs can be continuously changed when $\alpha>1$ or $\alpha<0$.
Thus, it is expected that the rule dependency should get higher or lower when we set $\alpha>1$ or $\alpha<0$, respectively.
We find that the rule is effective for \textsc{Target 1} and the error is monotonically decreased until $\alpha=1$ when we apply the trained model on \textsc{Source} (Fig.~\ref{fig:cardio-da-target1}).
The decreasing-trend of the error is continued as $\alpha$ is increased until 1.4 and it extends the range of $\alpha$ for further controllability.
This observation is particularly relevant when it is necessary to increase the rule dependency more and underlines how {\ours} learns rule representations effectively.

\section{Ablation Studies}\label{sec:ablation}
\vspace{-0.05in}

\begin{figure}[t]
    \centering
    \begin{subfigure}[b]{.43\linewidth}
        \centering
        \includegraphics[width=\textwidth]{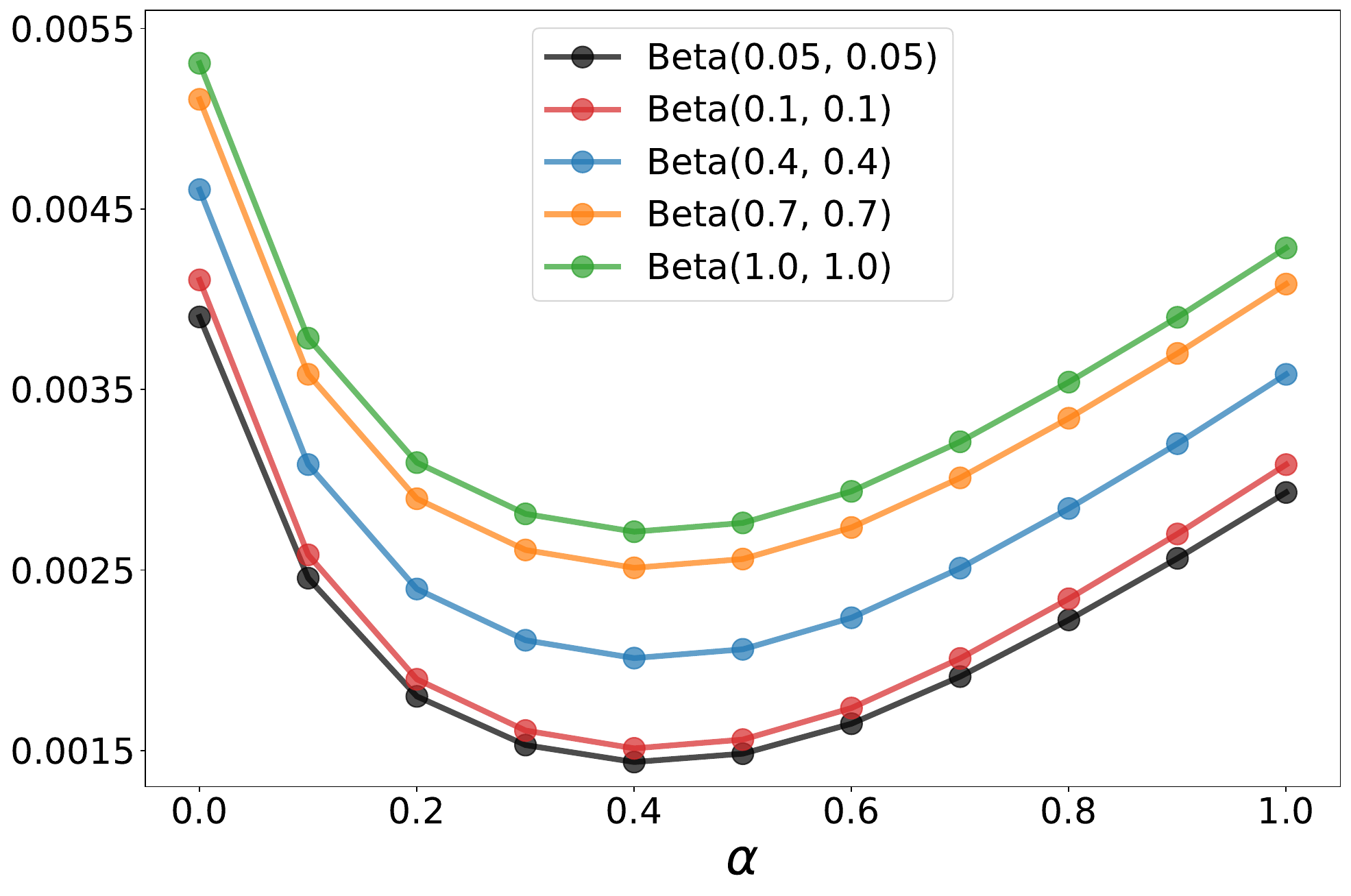}
        \caption{MAE}\label{fig:dp-mae-beta}
    \end{subfigure}
    \begin{subfigure}[b]{.408\linewidth}
        \centering
        \includegraphics[width=\textwidth]{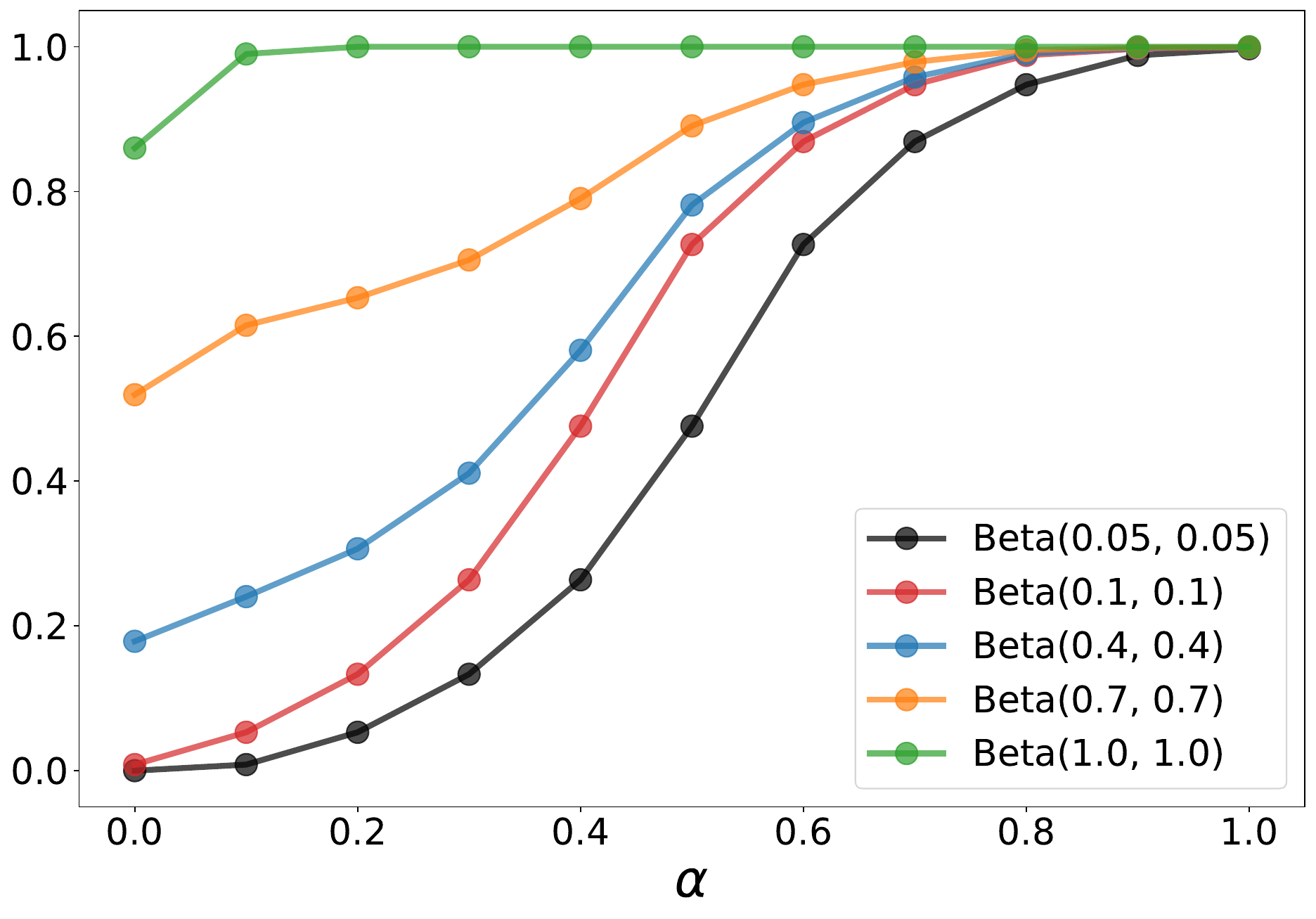}
        \caption{Verification ratio}\label{fig:dp-ratio-beta}
    \end{subfigure}
    \begin{subfigure}[b]{.43\linewidth}
        \centering
        \includegraphics[width=\textwidth]{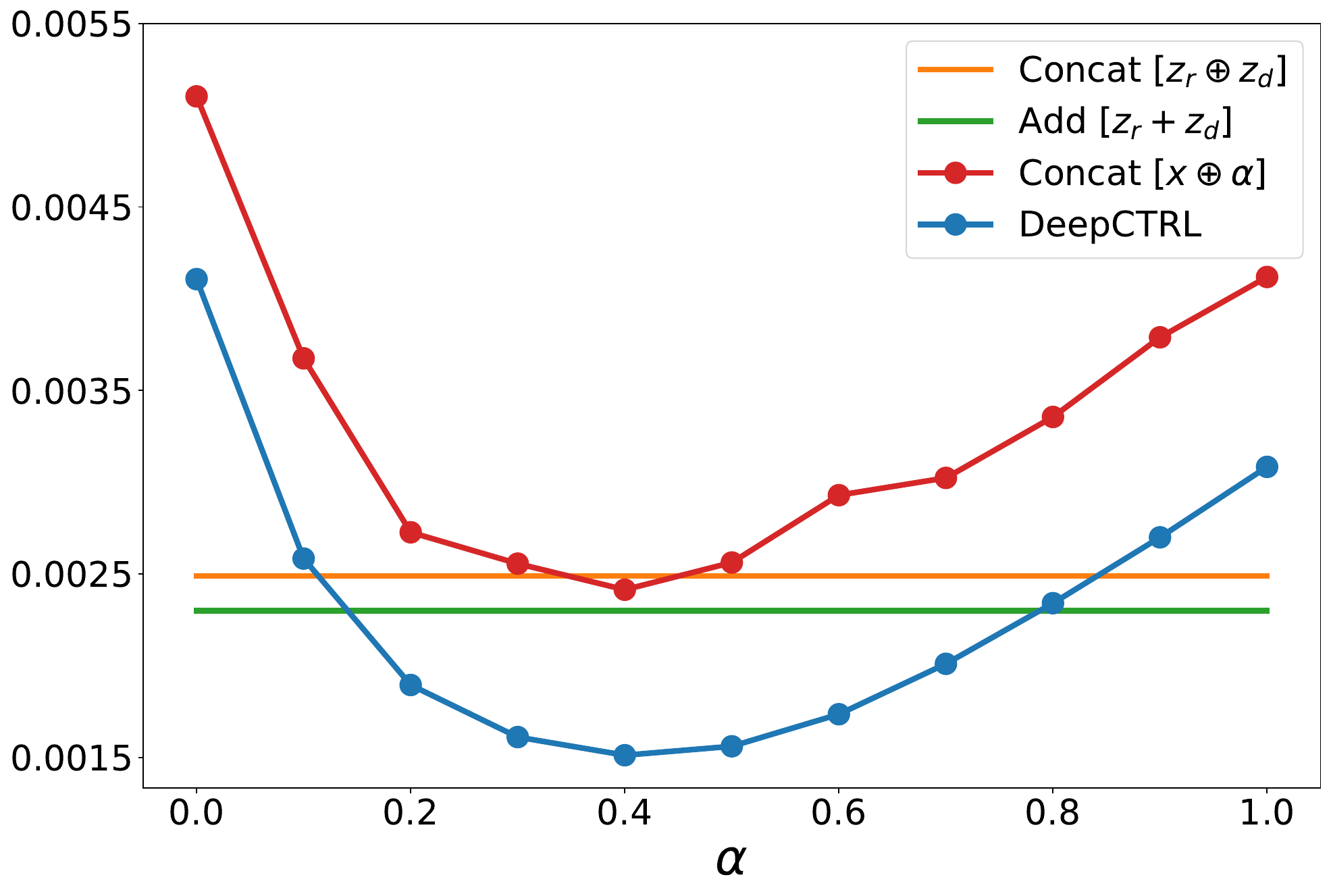}
        \caption{MAE}\label{fig:dp-mae-concat-add}
    \end{subfigure}
    \begin{subfigure}[b]{.408\linewidth}
        \centering
        \includegraphics[width=\textwidth]{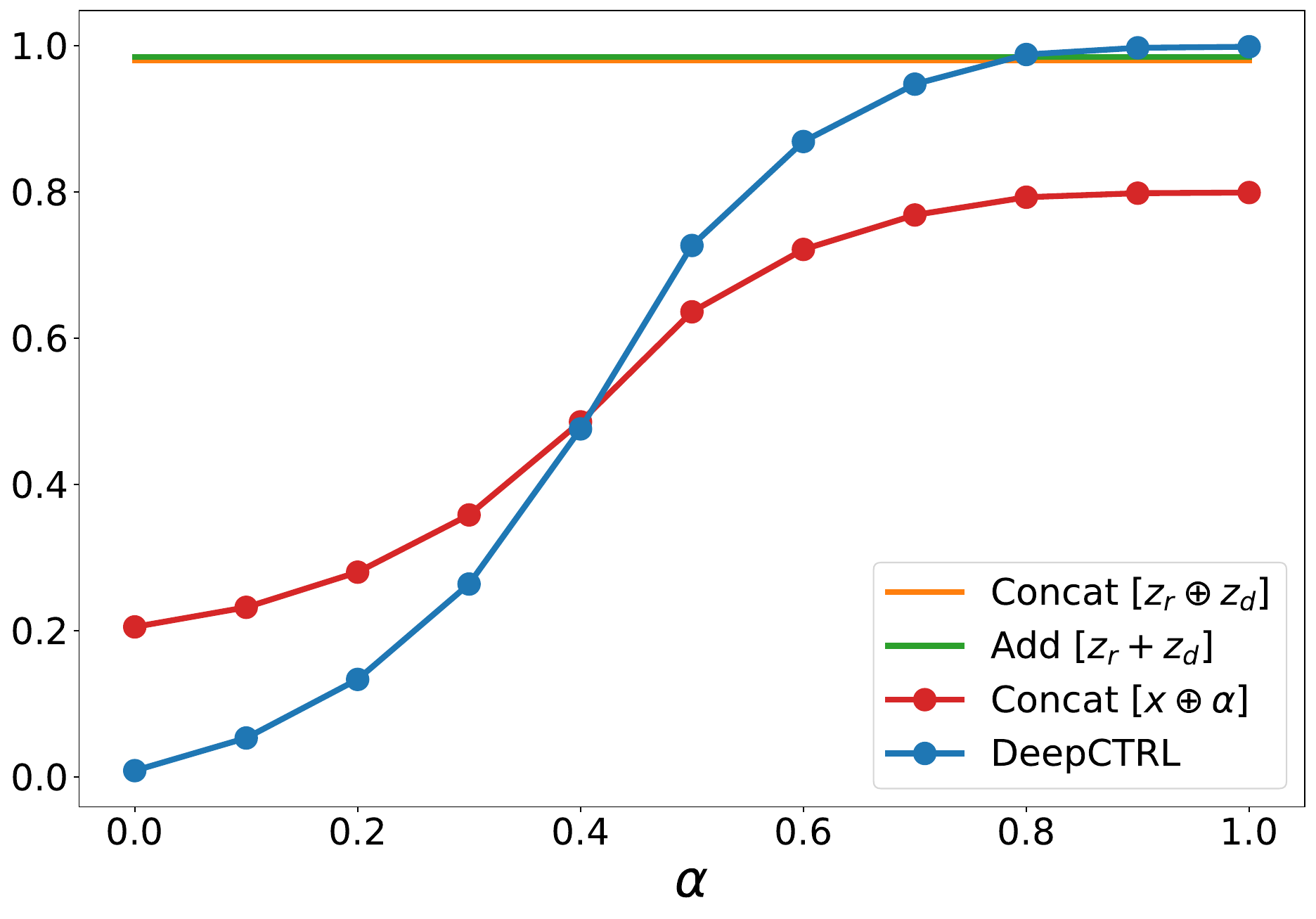}
        \caption{Verification ratio}\label{fig:dp-ratio-concat-add}
    \end{subfigure}
    \caption{(a,b) Results on a double pendulum task for various of $\beta$. (c,d) Comparison with two non-coupled representations (\textsc{Concat} and \textsc{Add}) and a concatenated input $(\vx\oplus\alpha)$.} \label{fig:dp-abblation}
\end{figure}

\textbf{Rule strength prior $P(\alpha)$:}
%
To ensure that {\ours} yields desired behavior with a range of $\alpha$ values at inference, we learn to mimic predictions with different $\alpha$ during training to provide the appropriate supervision. 
We choose the Beta distribution to sample $\alpha$ because via only one parameter, it allows us to sweep between various different behaviors.
One straightforward idea is to uniformly simulate all potential inference scenarios during training (i.e. having Beta(1,1)), however, this yields empirically worse results.
We propose that overemphasizing on edge cases (very low and very high $\alpha$) is important so that the behavior for edge cases can be learned more robustly, and the interpolation between them would be accurate.
Fig.~\ref{fig:dp-mae-beta} and~\ref{fig:dp-ratio-beta} shows results with different values of Beta($\beta,\beta$). 
For the pendulum task, $\gL_{rule}$ is much larger than $\gL_{task}$ and it leads {\ours} getting mostly affected by $\gL_{rule}$ on Beta$(1.0,1.0)$ when $\alpha>0.1$, hurting the task-specific performance.
We observe that there is typically an intermediate $\beta < 1$ optimal for the general behavior of the curves -- $\beta=0.1$ seems to be a reasonable choice across all datasets, and the results are not sensitive when it is slightly higher or lower. 

\textbf{Stochastically-coupled representations:}
We propose to combine the two representations from the encoders to enable stochastic coupling with corresponding objectives ($\alpha\vz_r, \alpha\gL_{rule}$ and $(1-\alpha)\vz_d, (1-\alpha)\gL_{task}$).
As an alternative, we consider the coupled representations with two conventional ways of merging: \textsc{Concat} ($\vz=\vz_r\oplus\vz_d$) and \textsc{Add} ($\vz=\vz_r+\vz_d$).
Fig.~\ref{fig:dp-mae-concat-add} and~\ref{fig:dp-ratio-concat-add} shows that these can provide a very high verification ratio (as $\gL_{rule}$ dominates over $\gL_{task}$), equivalent to {\ours} at high $\alpha$, but their errors at intermediate $\alpha$ values are higher than {\ours}.
The proposed coupling method in {\ours} enables disentangled learning of rule vs. task representations by mixing them with $\alpha$.
We also concatenate $\alpha$ as $[x;\alpha]$ instead of relying on the proposed two-passage network architecture.
Fig.~\ref{fig:dp-mae-concat-add} and~\ref{fig:dp-ratio-concat-add} shows that concatenating inputs does not improve the task error or verification ratio, although the same amount of information is used.
This supports the hypothesis that the coupled separation ($\bm{z}_r$ and $\bm{z}_d$) is more desirable for controlling corresponding representations.






\section{Conclusion and Societal Impact}
\label{sec:conclusion}
\vspace{-0.05in}

Learning from rules can be crucial for constructing interpretable, robust, and reliable DNNs.
In this paper, we propose {\ours}, a new methodology to incorporate rules into data-learned DNNs.
Unlike existing methods, {\ours} enables controllability of rule strength at inference without retraining. 
We propose a novel perturbation-based rule encoding method to integrate arbitrary rules into meaningful representations.
Overall, {\ours} is model architecture, data type and rule type agnostic. 
We demonstrate three uses cases of {\ours}: improving reliability given known principles, examining candidate rules, and domain adaptation using the rule strength. 
We leave theoretical proofs for convergence of learning, extensive empirical performance analysis on other datasets, as well as demonstrations of other use cases like robustness, for future work.

{\ours} has many potential benefits in real-world deep learning deployments to improve their accuracy, to increase their reliability, and to enhance human-AI interaction. On the other hand, we also note the capability of {\ours} in encoding rules in effective ways can have undesired outcomes if used with bad intentions to teach unethical biases.




\newpage
\small

\begin{thebibliography}{10}

\bibitem{arik2020protoattend}
Sercan~O Ar{\i}k and Tomas Pfister.
\newblock Protoattend: Attention-based prototypical learning.
\newblock {\em Journal of Machine Learning Research}, 21:1--35, 2020.

\bibitem{tabnet}
Sercan~{\"{O}}mer Arik and Tomas Pfister.
\newblock Tabnet: Attentive interpretable tabular learning.
\newblock {\em arXiv:1908.07442}, 2019.

\bibitem{arya2019one}
Vijay Arya, Rachel~KE Bellamy, Pin-Yu Chen, Amit Dhurandhar, Michael Hind,
  Samuel~C Hoffman, Stephanie Houde, Q~Vera Liao, Ronny Luss, Aleksandra
  Mojsilovi{\'c}, et~al.
\newblock One explanation does not fit all: A toolkit and taxonomy of ai
  explainability techniques.
\newblock {\em arXiv preprint arXiv:1909.03012}, 2019.

\bibitem{asseman2018learning}
Alexis Asseman, Tomasz Kornuta, and Ahmet Ozcan.
\newblock Learning beyond simulated physics.
\newblock In {\em Modeling and Decision-making in the Spatiotemporal Domain
  Workshop}, 2018.

\bibitem{breen2019newton}
Philip~G Breen, Christopher~N Foley, Tjarda Boekholt, and Simon~Portegies
  Zwart.
\newblock Newton vs the machine: solving the chaotic three-body problem using
  deep neural networks.
\newblock {\em arXiv preprint arXiv:1910.07291}, 2019.

\bibitem{Browning1986MicroeconomicTA}
Edgar~K. Browning and J.~M. Browning.
\newblock Microeconomic theory and applications.
\newblock 1986.

\bibitem{damour2020underspecification}
Alexander D'Amour, Katherine Heller, Dan Moldovan, Ben Adlam, Babak Alipanahi,
  Alex Beutel, Christina Chen, Jonathan Deaton, Jacob Eisenstein, Matthew~D.
  Hoffman, Farhad Hormozdiari, Neil Houlsby, Shaobo Hou, Ghassen Jerfel, Alan
  Karthikesalingam, Mario Lucic, Yian Ma, Cory McLean, Diana Mincu, Akinori
  Mitani, Andrea Montanari, Zachary Nado, Vivek Natarajan, Christopher Nielson,
  Thomas~F. Osborne, Rajiv Raman, Kim Ramasamy, Rory Sayres, Jessica Schrouff,
  Martin Seneviratne, Shannon Sequeira, Harini Suresh, Victor Veitch, Max
  Vladymyrov, Xuezhi Wang, Kellie Webster, Steve Yadlowsky, Taedong Yun,
  Xiaohua Zhai, and D.~Sculley.
\newblock Underspecification presents challenges for credibility in modern
  machine learning, 2020.

\bibitem{dash2018boolean}
Sanjeeb Dash, Oktay Gunluk, and Dennis Wei.
\newblock Boolean decision rules via column generation.
\newblock In {\em Advances in Neural Information Processing Systems},
  volume~31, pages 4655--4665. Curran Associates, Inc., 2018.

\bibitem{8260755}
Michelangelo Diligenti, Soumali Roychowdhury, and Marco Gori.
\newblock Integrating prior knowledge into deep learning.
\newblock In {\em 2017 16th IEEE International Conference on Machine Learning
  and Applications (ICMLA)}, pages 920--923, 2017.

\bibitem{doshi2017towards}
Finale Doshi-Velez and Been Kim.
\newblock Towards a rigorous science of interpretable machine learning.
\newblock {\em arXiv preprint arXiv:1702.08608}, 2017.

\bibitem{eisenach2020mqtransformer}
Carson Eisenach, Yagna Patel, and Dhruv Madeka.
\newblock Mqtransformer: Multi-horizon forecasts with context dependent and
  feedback-aware attention, 2020.

\bibitem{fioretto2020lagrangian}
Ferdinando Fioretto, Pascal~Van Hentenryck, Terrence~WK Mak, Cuong Tran,
  Federico Baldo, and Michele Lombardi.
\newblock Lagrangian duality for constrained deep learning, 2020.

\bibitem{posteriorreg}
K.~Ganchev, J.~Graça, J.~Gillenwater, and B.~Taskar.
\newblock Posterior regularization for structured latent variable models.
\newblock {\em J. Mach. Learn. Res.}, 11:2001--2049, 2010.

\bibitem{Goodfellow-et-al-2016}
Ian Goodfellow, Yoshua Bengio, and Aaron Courville.
\newblock {\em Deep Learning}.
\newblock MIT Press, 2016.
\newblock \url{http://www.deeplearningbook.org}.

\bibitem{goodfellow2014explaining}
Ian~J Goodfellow, Jonathon Shlens, and Christian Szegedy.
\newblock Explaining and harnessing adversarial examples.
\newblock {\em arXiv preprint arXiv:1412.6572}, 2014.

\bibitem{greydanus2019hamiltonian}
Samuel Greydanus, Misko Dzamba, and Jason Yosinski.
\newblock Hamiltonian neural networks.
\newblock In {\em Advances in Neural Information Processing Systems}, pages
  15379--15389, 2019.

\bibitem{hestness2017deep}
Joel Hestness, Sharan Narang, Newsha Ardalani, Gregory Diamos, Heewoo Jun,
  Hassan Kianinejad, Md. Mostofa~Ali Patwary, Yang Yang, and Yanqi Zhou.
\newblock Deep learning scaling is predictable, empirically, 2017.

\bibitem{harnessdnnlogicrules}
Zhiting Hu, Xuezhe Ma, Zhengzhong Liu, Eduard~H. Hovy, and Eric~P. Xing.
\newblock Harnessing deep neural networks with logic rules.
\newblock {\em CoRR}, abs/1603.06318, 2016.

\bibitem{KANNEL2000251}
William~B Kannel.
\newblock Elevated systolic blood pressure as a cardiovascular risk factor.
\newblock {\em The American Journal of Cardiology}, 85(2):251--255, 2000.

\bibitem{kurakin2016adversarial}
Alexey Kurakin, Ian Goodfellow, and Samy Bengio.
\newblock Adversarial examples in the physical world.
\newblock {\em arXiv preprint arXiv:1607.02533}, 2016.

\bibitem{tft}
Bryan Lim, Sercan~O. Arik, Nicolas Loeff, and Tomas Pfister.
\newblock Temporal fusion transformers for interpretable multi-horizon time
  series forecasting, 2020.

\bibitem{liu2020very}
Xiaodong Liu, Kevin Duh, Liyuan Liu, and Jianfeng Gao.
\newblock Very deep transformers for neural machine translation.
\newblock {\em arXiv preprint arXiv:2008.07772}, 2020.

\bibitem{lutter2019deep}
Michael Lutter, Christian Ritter, and Jan Peters.
\newblock Deep lagrangian networks: Using physics as model prior for deep
  learning, 2019.

\bibitem{pmlr-v84-narasimhan18a}
Harikrishna Narasimhan.
\newblock Learning with complex loss functions and constraints.
\newblock In Amos Storkey and Fernando Perez-Cruz, editors, {\em Proceedings of
  the Twenty-First International Conference on Artificial Intelligence and
  Statistics}, volume~84 of {\em Proceedings of Machine Learning Research},
  pages 1646--1654. PMLR, 09--11 Apr 2018.

\bibitem{node}
Sergei Popov, Stanislav Morozov, and Artem Babenko.
\newblock Neural oblivious decision ensembles for deep learning on tabular
  data.
\newblock {\em arXiv:1909.06312}, 2019.

\bibitem{ribeiro2016should}
Marco~Tulio Ribeiro, Sameer Singh, and Carlos Guestrin.
\newblock " why should i trust you?" explaining the predictions of any
  classifier.
\newblock In {\em Proceedings of the 22nd ACM SIGKDD international conference
  on knowledge discovery and data mining}, pages 1135--1144, 2016.

\bibitem{shao2020controlvae}
Huajie Shao, Shuochao Yao, Dachun Sun, Aston Zhang, Shengzhong Liu, Dongxin
  Liu, Jun Wang, and Tarek Abdelzaher.
\newblock Controlvae: Controllable variational autoencoder.
\newblock In {\em International Conference on Machine Learning}, pages
  8655--8664. PMLR, 2020.

\bibitem{szegedy2016rethinking}
Christian Szegedy, Vincent Vanhoucke, Sergey Ioffe, Jon Shlens, and Zbigniew
  Wojna.
\newblock Rethinking the inception architecture for computer vision.
\newblock In {\em Proceedings of the IEEE conference on computer vision and
  pattern recognition}, pages 2818--2826, 2016.

\bibitem{touvron2019fixing}
Hugo Touvron, Andrea Vedaldi, Matthijs Douze, and Herve Jegou.
\newblock Fixing the train-test resolution discrepancy.
\newblock In {\em Advances in Neural Information Processing Systems}, pages
  8252--8262, 2019.

\bibitem{vaswani2017attention}
Ashish Vaswani, Noam Shazeer, Niki Parmar, Jakob Uszkoreit, Llion Jones,
  Aidan~N Gomez, {\L}ukasz Kaiser, and Illia Polosukhin.
\newblock Attention is all you need.
\newblock In {\em Advances in neural information processing systems}, pages
  5998--6008, 2017.

\bibitem{yuan2019adversarial}
Xiaoyong Yuan, Pan He, Qile Zhu, and Xiaolin Li.
\newblock Adversarial examples: Attacks and defenses for deep learning.
\newblock {\em IEEE transactions on neural networks and learning systems},
  30(9):2805--2824, 2019.

\bibitem{adversarialdebiasing}
Brian~Hu Zhang, Blake Lemoine, and Margaret Mitchell.
\newblock Mitigating unwanted biases with adversarial learning.
\newblock {\em CoRR}, abs/1801.07593, 2018.

\bibitem{zhang2018mixup}
Hongyi Zhang, Moustapha Cisse, Yann~N. Dauphin, and David Lopez-Paz.
\newblock mixup: Beyond empirical risk minimization.
\newblock In {\em International Conference on Learning Representations}, 2018.

\end{thebibliography}

\newpage
\appendix


\section{Appendix}

\subsection{Experimental Settings}
We provide the detailed model configurations and hyperparameter settings for {\ours} in the following three experiments: double pendulum dynamics, sales forecasting, and cardiovascular classification.

\subsubsection{Model Configurations}

\paragraph{Double Pendulum}
The input and output states are 4 dimensional (two angular displacements and two angular velocities).
The input state is fed into a shared layer whose configuration is {[FC64,ReLU,FC16]} where {FC($n$)} denotes a fully-connected layer with $n$ units.
Then, output from the shared layer is fed into two encoders: Rule encoder {[FC64,ReLU,FC64,ReLU,FC64]} and Data encoder {[FC64,ReLU,FC64,ReLU,FC64]}.
The combined representation from the two encoders is fed into a decision block {[FC64,ReLU,FC4]}.
We use Adam optimization algorithm for training with learning rate 0.001.

\paragraph{Sales Forecasting}
The input dimension is 13, consisting of an item price as well as derivative features, and the output dimension is 1 (total weekly sales).
Both encoders have same configuration {[FC64,ReLU,FC64,ReLU,FC16]} followed by a decision block {[FC64,ReLU,FC1]}.
We use Adam optimization algorithm for training with learning rate 0.001.

\paragraph{Cardiovascular Classification}
The number of original input features is 11: \textsc{Age, Height, Weight, Gender, Systolic blood pressure, Diastolic blood pressure, Cholesterol, Glucose, Smoking, Alcohol intake, Physical activity}.
We expand categorical features to one-hot encoding (the input dimension is increased to 19) and the output dimension is 1 (Presence or absence of cardiovascular disease).
Both encoders have same configuration {[FC100,ReLU,FC16]} followed by a decision block {[FC1,Sigmoid]}.
We use Adam optimization algorithm for training with learning rate 0.001.


\subsubsection{Data Splits}
\paragraph{Double Pendulum}
The total length of simulated dynamics is 30,000 and it is split into training (18,000), validation (3,000), and testing (9,000).

\paragraph{Sales Forecasting}
Per the Kaggle task definition, first 273 weeks are used in a training set, the following 4 weeks are for a validation set, and the last 4 weeks are for a testing set.
In each set, there are 22,543, 700, and 700 records from the preprocessed dataset.

\paragraph{Cardiovascular Classification}
The data partitioning is described in Section 4.3.
For \textsc{Source} partition, we have 70\% training samples, 10\% validation samples, and 20\% testing samples to train and evaluate a model.
Then, the trained model is applied to \textsc{Target 1, Target 2}, and \textsc{Target 3}, respectively.

\setlength{\tabcolsep}{3pt}
\begin{table}[!htbp]
\caption{Dataset splits and \textsc{Usual} vs. \textsc{Unusual} partitioning.}
\label{tab:da}
\begin{center}
\begin{small}
\begin{sc}
\begin{tabular}{lcccc}
\toprule
Data & Source & Target 1 & Target 2 & Target 3 \\
\midrule
\textsc{Usual}   & 6,007 & 20,000 & 6,000 & 4,000 \\
\textsc{Unusual} & 14,018 & 6,009 & 6,009 & 6,009 \\
Ratio & 0.30 & 0.77 & 0.50 & 0.40 \\
\bottomrule
\end{tabular}
\end{sc}
\end{small}
\end{center}
\end{table}

\subsubsection{Training Settings}
We commonly train {\ours} for all tasks with a batch size of 32 on a single GPU (Nvidia T4 GPU) for 1000 epochs with early stopping where a validation error is not improved for 10 epochs.
All results in the paper are mean values from 10 different random seeds.

\subsection{Perturbations}

\subsubsection{How to set \texorpdfstring{$\delta\vx$}{dx}?}
In Section 3, we provide a method to integrate non-differentiable $\gL_{rule}$ via input perturbations.
In this paper, we use the rules obtained via perturbation-based method for two tasks: sales forecasting (Section 4.2) and cardiovascular classification (Section 4.3).
We summarize how the perturbations are generated and used to define the corresponding rule-based constraint for each task.
Note that both tasks have a similar non-differentiable form of rule-constraint that the input $\vx$ and output $\vy$ have a negative or positive correlation.

\paragraph{Perturbations in Sales Forecasting}
There are a number of weekly-sales-records per an item at a particular store. 
For each record (week $t$), we have input features including the price of an item $\vx_t$ and the target sales $\vy_t$.
The correlation coefficient between $\vx$ and $\vy$ is: 
\begin{align}
    R = \frac{\sum_{t=1}^T(\vx_t-\overline{\vx})(\vy_t-\overline{\vy})}{\sqrt{\sum_{t=1}^T(\vx_t-\overline{\vx})^2}\sqrt{\sum_{t=1}^T(\vy_t-\overline{\vy})^2}}, \label{app-eq:corr}
\end{align}
where $\overline{\vx}$ and $\overline{\vy}$ are the sample mean of $\vx_t$ and $\vy_t$, respectively.

The rule-based constraint we want to impose is \textit{price and sales should have a negative correlation coefficient}, i.e. $R<0$.
While the constraint is based on the exact definition of the correlation coefficient over $T$ samples (Eq.~\ref{app-eq:corr}), we use a constraint based on individual sample instead:
\begin{align}
    \frac{\Delta\vy}{\Delta\vx}=\frac{\vy_p - \vy}{\vx_p - \vx}< 0, \label{app-eq:delta-corr}
\end{align}
where $\Delta\vx$ is a price-difference and $\Delta\vy$ is a sales-difference, respectively, and $\vx_p$ is a perturbed price where $\vx_p=\vx+\delta\vx$ and $\vy_p$ is an output from the perturbed price.
Note that Eq.~\ref{app-eq:delta-corr} is identical to $\vy_p<\vy$ once $\delta\vx>0$.
There are two reasons not to use Eq.~\ref{app-eq:corr} directly.
First, it is costly to compute the coefficient at every iteration.
Second, it is possible to control $\delta\vx$ in Eq.~\ref{app-eq:delta-corr}.
We set $\delta\vx=\gamma|\vx|$, where $\gamma\sim\textsc{U}[0,u]$ such that $\gamma$ is a perturbation scale parameter and $u$ is an upper bound of $\gamma$.
Thus, the magnitude of $\delta\vx$ is bounded by $[0, u|\vx|)$.

\paragraph{Perturbations in Cardiovascular Classification}
Similarly, we impose a positive correlation between blood pressure $\vx$ and a risk of the cardiovascular disease $\vy$:
\begin{align}
    \frac{\Delta\vy}{\Delta\vx}=\frac{\vy_p - \vy}{\vx_p - \vx}> 0. \label{app-eq:delta-corr-cardio}
\end{align}
Eq.~\ref{app-eq:delta-corr-cardio} is analogous to $\vy_p>\vy$ as long as the perturbation $\delta\vx>0$.

\paragraph{Upper Bound $u$:}
We set the upper bound $u$ as 0.1 for both tasks via the analysis described in Section~\ref{app-sec:impact-of-pert}.

\subsubsection{Impact of Perturbation Scale} \label{app-sec:impact-of-pert}

\begin{figure}[t]
    \centering
    \begin{subfigure}[b]{.475\linewidth}
        \centering
        \includegraphics[width=\textwidth]{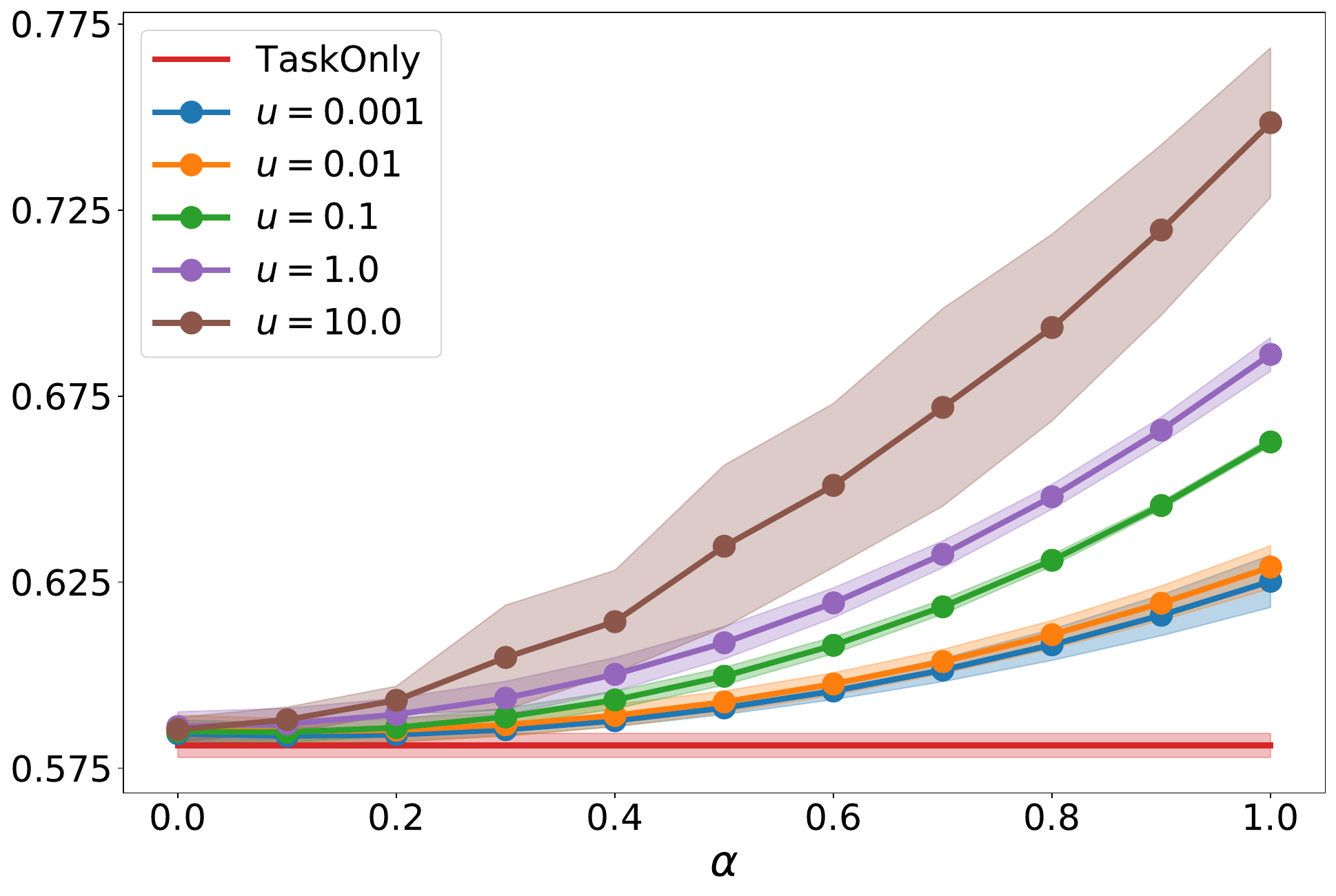}
        \caption{Cross Entropy}\label{app-fig:cardio-pert-ce}
    \end{subfigure}
    \hfill
    \begin{subfigure}[b]{.475\linewidth}
        \centering
        \includegraphics[width=\textwidth]{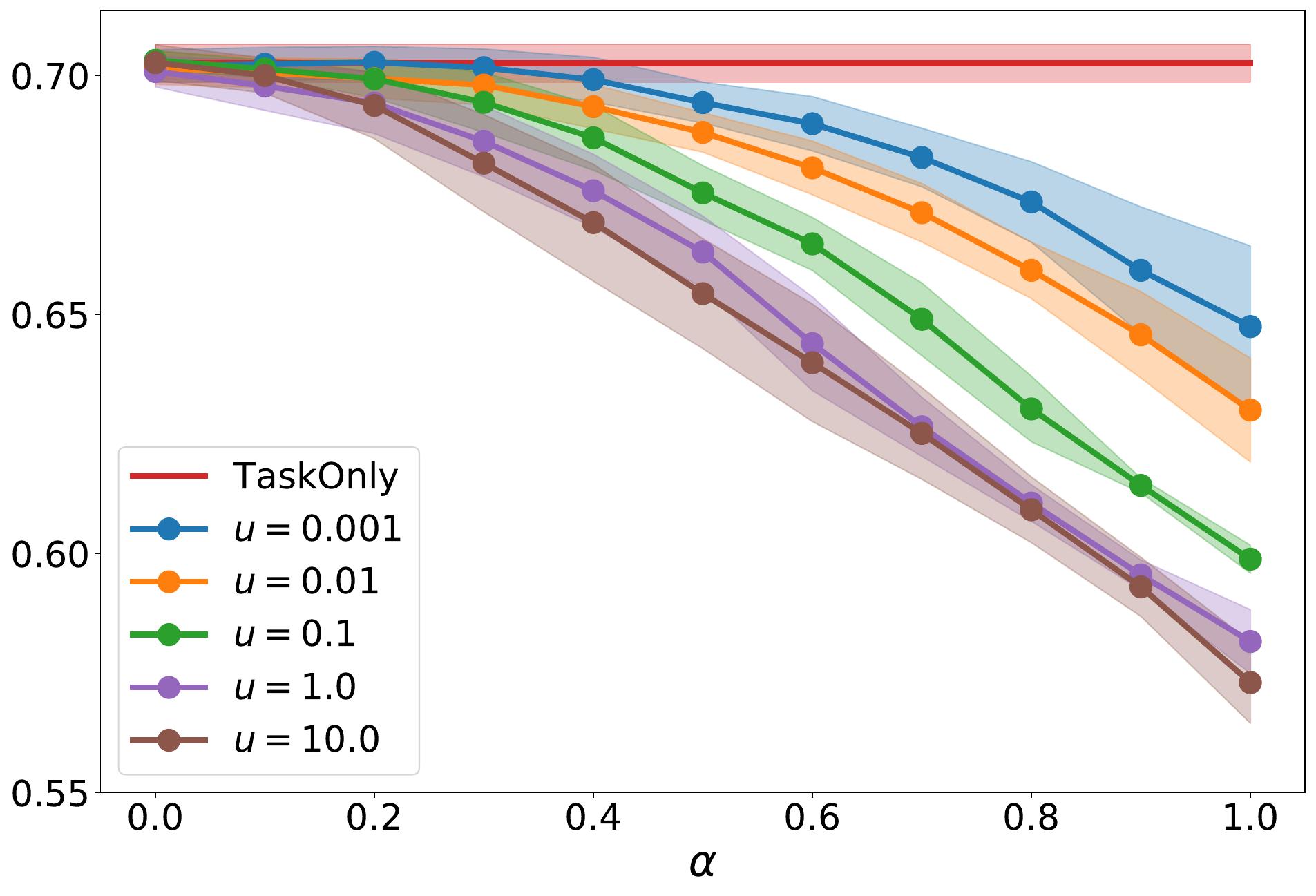}
        \caption{Accuracy}\label{app-fig:cardio-pert-acc}
    \end{subfigure}
    \caption{(Left) Cross entropy and (Right) Accuracy vs. rule strength for various upper bounds of perturbation scale.} \label{app-fig:cardio-pert}
    \vspace{-1em}
\end{figure}

In previous section, we define $\delta\vx$ as a random scalar that is upper-bounded by $u|\vx|$.
We experiment on the cardiovascular classification task with different upper bounds, $u$ to see how the scale of perturbation affects the model's behavior.

Fig~\ref{app-fig:cardio-pert} shows how the cross entropy and accuracy are changed as the rule strength is changed.
When the upper bound $u$ is increased, the perturbation scale $\gamma$ can be also increased, and thus, it leads to generate larger perturbations $\delta\vx$.
As $u$ increases, the performance of a classifier is degraded when rule strength is non zero ($\alpha>0$).
The curves imply that larger $u$ leads more degraded performance when the rule strength is higher.
When $\delta\vx$ increases, the perturbed output $\vy_p$ is more different to $\vy$ as $\vy_p$ is a function of $\vx+\delta\vx$.
According to Eq.~\ref{app-eq:delta-corr} and~\ref{app-eq:delta-corr-cardio}, the rule-based objective $\gL_{rule}$ is a function of $\vy_p - \vy$, and thus, the larger $\delta\vx$ eventually causes larger $\gL_{rule}$.
In other words, as $\gL_{rule}$ increases, {\ours} is more driven by the rule-based objective when $\alpha$ is non zero and thus, the task-based performance is degraded.
As discussed, it is desired to have distinct model's behavior when $\alpha=0$ and $\alpha=1$ and thus, too small perturbation less incorporate rule-based representations.
However, if too large perturbations are considered, the model is dominated by the rule mostly and the performance can be worse when $\alpha$ is close to 0 (the brown curve is slightly higher than others when $\alpha\rightarrow0$ in Fig.~\ref{app-fig:cardio-pert-ce}).

\subsection{Visualization of \texorpdfstring{$z_r,z_d$}{zrzd} and \texorpdfstring{$z$}{z}}

\begin{figure*}[t]
    \centering
    \begin{subfigure}[b]{.45\textwidth}
        \centering
        \includegraphics[width=0.95\textwidth]{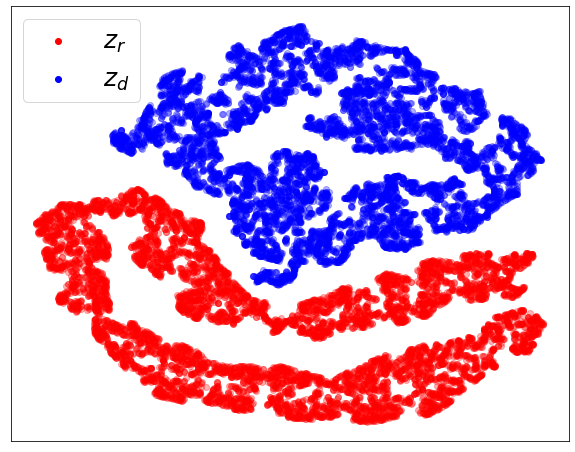}
        \caption{t-SNE mapping of $z_r$ and $z_d$.}\label{app-fig:encoder-embedding}
    \end{subfigure}
    \hfill
    \begin{subfigure}[b]{.45\textwidth}
        \centering
        \includegraphics[width=0.95\textwidth]{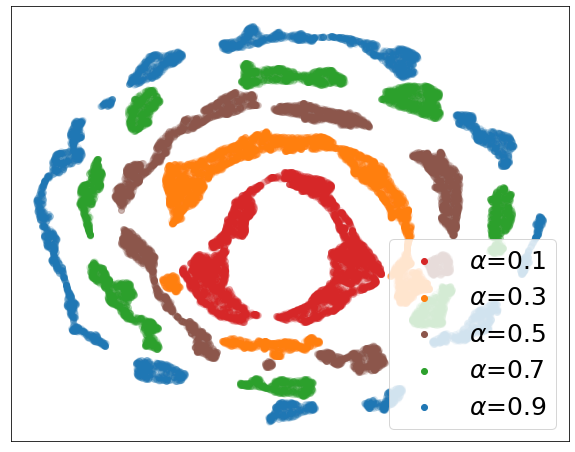}
        \caption{t-SNE mapping of $z=\alpha z_r+(1-\alpha z_d)$ over different $\alpha$.}\label{app-fig:z-emb}
    \end{subfigure}
    \caption{(Left) t-SNE visualization of $z_r$ and $z_d$ from rule encoder $\phi_r$ and data encoder $\phi_d$, respectively, from the double pendulum task. $z_r$ and $z_d$ from 9,000 test samples do not have overlapped representations and it implies that two representations are distinct. (Right) t-SNE visualization of $z=\alpha z_r + (1-\alpha)z_d$ over different $\alpha$. It shows that task-/rule-specific representations are placed in inner/outer space, respectively. Note that the proposed representations gradually interpolate two extremes rather than being abruptly crossed.}\label{app-fig:tsne}
    \vspace{-1em}
\end{figure*}

In this section, we analyze the learned representations to demonstrate the rule vs. data disentanglement capability of {\ours}.
We first visualize what the rule encoder $\phi_r$ and data encoder $\phi_d$ learn to support that the two encoders actually handle distinct representations rather than similar or overlapped representations.
Then, we show how $\vz=\alpha\vz_r + (1-\alpha)\vz_d$ is changed over different $\alpha$ to show meaningful combined representations with varying rule strength.
Fig.~\ref{app-fig:tsne} demonstrates t-SNE mapping of $\vz_r,\vz_d$, and $\vz$, respectively. The learned representations of $\vz_r$ and $\vz_d$ are observed to be separated, while their linear combinations are clustered together with an orientation of the clusters highly dependent on the rule strength.

\subsection{Computational Complexity}
Compared to {\taskonly} training, the proposed method (\ours) does not cause any additional computations that are proportional to the sample size.
However, if perturbations are included, teaching rules via perturbation-based method cause an additional computation.
For each batch, it is required to compute both the main forward pass ($\vx$ to $\vy$) and the perturbation-based forward pass ($\vx_p$ to $\vy_p$).
Thus, the time complexity of {\ours} is $2C_{forward} + C_{backward}$ and that of {\taskonly} is $C_{forward} + C_{backward}$ where $C_{forward}$ is the time complexity of forward propagation and $C_{backward}$ of backward propagation.
As $C_{backward}$ takes the majority of training time, the extra computations from the perturbation-based method is not significant.
Furthermore, since the computational complexity is independent on the size of samples, {\ours} is still scalable. Table~\ref{app-tab:running-time} shows the training time from {\dataonly} and {\ours} over 10 repeats.
Note that the running time per an epoch from {\ours} is similar to that of {\dataonly} and it proves that the computational complexity of {\ours} is not significantly increased.

\begin{table}[t]
\centering
\caption{Training time (in seconds) for Sales forecasting (Retail) and Cardiovascular classification (Healthcare).}
\label{app-tab:running-time}
\begin{center}
\begin{small}
\begin{sc}
\begin{tabular}{lcccc}
\toprule
Data & {\dataonly} (total) & {\ours} (total) & {\dataonly} (per epoch) & {\ours} (per epoch) \\
\midrule
Retail  & 362.7$\pm$62.3 & 328.2$\pm$91.0 & 2.31$\pm$0.04 & 2.32$\pm$0.01 \\
Healthcare  & 154.7$\pm$11.0 & 168.1$\pm$23.1 & 3.33$\pm$0.24 & 3.45$\pm$0.17\\
\bottomrule
\end{tabular}
\end{sc}
\end{small}
\end{center}
\end{table}

\subsection{Adaptive loss combination:}

\begin{figure}[t]
    \centering
    \begin{subfigure}[b]{.475\linewidth}
        \centering
        \includegraphics[width=\textwidth]{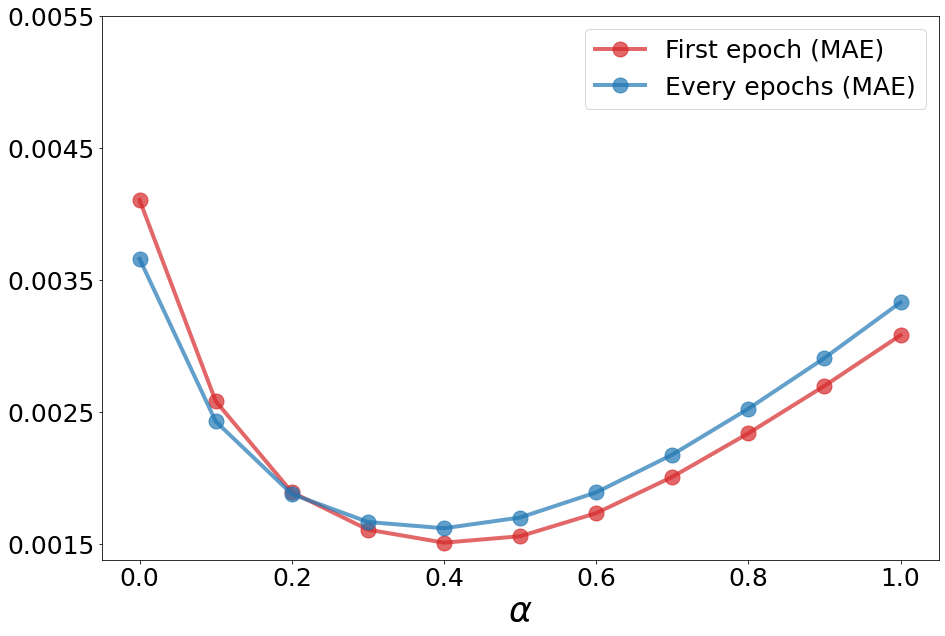}
        \caption{MAE}
    \end{subfigure}
    \hfill
    \begin{subfigure}[b]{.46\linewidth}
        \centering
        \includegraphics[width=\textwidth]{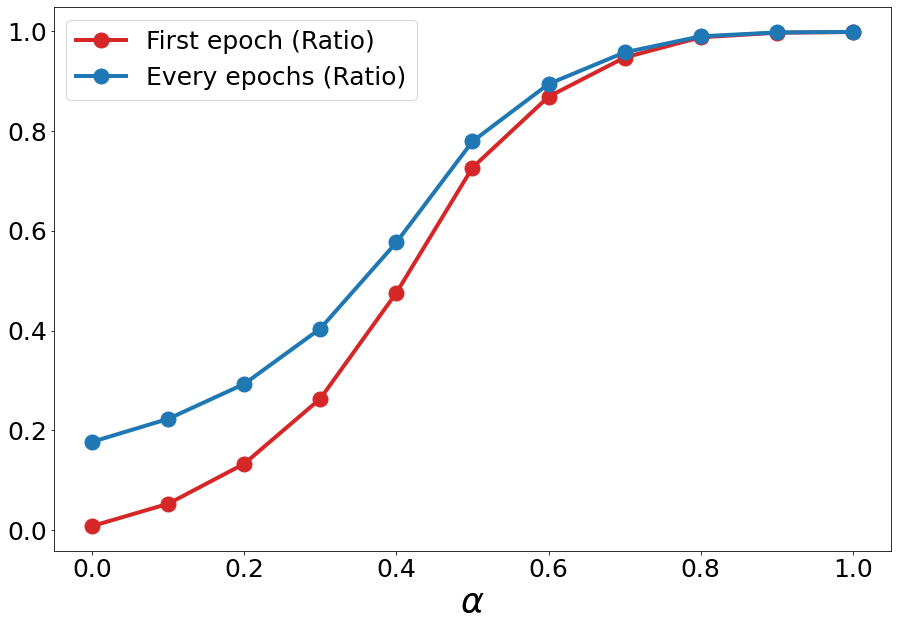}
        \caption{Verification ratio}
    \end{subfigure}
    \caption{Results on a double pendulum task (different $\rho$).} \label{fig:author-dp-rho}
    \vspace{-2em}
\end{figure}

To balance the contributions from $\gL_{task}$ and $\gL_{rule}$, we propose to use the scale parameter $\rho=\gL_{rule,0}/\gL_{task,0}$. One potential issue with this proposal is that at the beginning of training, the model is far from convergence, and the initial estimates of loss values may not be representative. One straightforward idea to address could be adapting $\rho$ as the ratio after every epoch. In Fig. \ref{fig:author-dp-rho}, we compare this approach to the proposed $\rho=\gL_{rule,0}/\gL_{task,0}$ and we indeed show that the results are quite similar, and indeed $\rho=\gL_{rule,0}/\gL_{task,0}$ yields better decoupling of $\alpha=0$ cases (as evident from 0 verification ratio).

We attribute this to the fact that the scale mismatching mostly happens at early phase of training, and $\mathcal{L}_{rule}$ and $\mathcal{L}_{task}$ become rapidly very small and the parameters are not significantly changed. This property is dependent on the type of rule, type of task, and dataset so that it is necessary to choose a proper method beforehand.
Having a fixed coefficient is particularly beneficial for stable training because the model has a fixed target, rather than a varying one.

\end{document}